\documentclass[12pt]{article}
\usepackage{amsmath}
\usepackage{times}
\usepackage{graphicx}
\usepackage{color}
\usepackage{multirow}
\usepackage{subcaption}
\usepackage{url}
\usepackage[authoryear]{natbib}
\usepackage{rotating}
\usepackage{bbm}
\usepackage{latexsym}
\usepackage{fullpage}
\usepackage{color}
\usepackage[linesnumbered,ruled,vlined]{algorithm2e}
\usepackage{array}
\newcolumntype{$}{>{\global\let\currentrowstyle\relax}}
\newcolumntype{^}{>{\currentrowstyle}}

\newcommand{\black}[1]{{\textcolor{black}{#1}}}

\DeclareMathOperator{\grad}{\nabla}
\DeclareMathOperator{\loss}{\mathcal{L}}
\DeclareMathOperator{\reg}{\mathcal{R}}
\DeclareMathOperator{\jacobian}{\mathcal{J}}

\newcommand{\captionfonts}{\normalsize}

\makeatletter  
\long\def\@makecaption#1#2{%
  \vskip\abovecaptionskip
  \sbox\@tempboxa{{\captionfonts #1: #2}}%
  \ifdim \wd\@tempboxa >\hsize
    {\captionfonts #1: #2\par}
  \else
    \hbox to\hsize{\hfil\box\@tempboxa\hfil}%
  \fi
  \vskip\belowcaptionskip}
\makeatother   

\begin{document}
\hspace{13.9cm}1

\ \vspace{20mm}\\

{\LARGE Unifying Adversarial Training Algorithms with Flexible Deep Data Gradient Regularization}

\ \\
{\bf \large Alexander G. Ororbia II$^{\displaystyle 1}$}\\
{\bf \large Daniel Kifer$^{\displaystyle 1}$}\\
{\bf \large C. Lee Giles$^{\displaystyle 1}$}\\
{$^{\displaystyle 1}$The Pennsylvania State University.}\\
%

{\bf Keywords:} neural architectures, adversarial examples, blind-spot problem, data-driven regularization, approximate deep gradient

\thispagestyle{empty}
\markboth{}{DataGrad}
\ \vspace{-0mm}\\
%
\begin{center} {\bf Abstract} \end{center}
Many previous proposals for adversarial training of deep neural nets have included directly modifying the gradient, training on a mix of original and adversarial examples, using contractive penalties, and approximately optimizing constrained adversarial objective functions. In this paper, we show these proposals are actually all instances of optimizing a general, regularized objective we call DataGrad. Our proposed DataGrad framework, which can be viewed as a deep extension of the layerwise contractive autoencoder penalty, cleanly simplifies prior work and easily allows extensions such as adversarial training with multi-task cues. In our experiments, we find that the deep gradient regularization of DataGrad (which also has L1 and L2 flavors of regularization) outperforms alternative forms of regularization, including classical L1, L2, and multi-task, both on the original dataset as well as on adversarial sets. Furthermore, we find that combining multi-task optimization with DataGrad adversarial training results in the most robust performance.

\section{Introduction}
Deep neural architectures are highly effective at a vast array of tasks, both supervised and unsupervised. However, recently, it has been shown that deep architectures are sensitive to certain kinds of pertubations of the input, which can range from being barely perceptible to quite noticeable (even semi-random noise), as in \cite{nguyen_deep_2014}. Samples containing this type of noise are called ``adversarial examples'' \citep{szegedy_intriguing_2013} and can cause a trained network to confidently misclassify its input.  While there are a variety of ways to generate adversarial samples, the fastest and most effective approaches in the current literature are based on the idea of using back-propagation to acquire the derivative of the loss with respect to an input image (i.e. the gradient) and adding a small multiple of the gradient to the image.

Earlier work suggested adding a regularization penalty on the deep gradient \citep{goodfellow_explaining_2014,gu_towards_2014}, but had difficulty in computing the derivative (with respect to the weights) of the gradient, which is necessary for gradient-descent based algorithms. Instead, they utilized approximations such as a shallow layerwise gradient penalty \citep{gu_towards_2014}, also used for regularizing contractive auto-encoders \citep{gu_towards_2014}. \black{Meanwhile \citep{lyuunify} also presented a heuristic algorithm for this objective.}

Here we provide an efficient, deterministic back-propagation style algorithm for training with a wide variety of gradient (or rather, gradient) penalties.  It would seem that the resulting algorithm has the potential for unifying existing approaches for deep adversarial training. In particular, it helps explain some of the newer approaches to adversarial training \citep{miyato_distributional_2015,huang_learning_2015}. These approaches set up an adversarial objective as a constrained optimization problem and then approximate/simplify it using properties that hold for optimal solutions of \emph{unconstrained} problems. The algorithms then developed approximate optimization (when compared to our algorithms) and can be viewed as regularizations of this deep gradient.



\section{The DataGrad Framework}
\label{unified_view}
Given a set of loss functions $\loss_0,\loss_1,\dots,\loss_m$ and regularizers $\reg_1,\dots,\reg_m$, consider:
\begin{align*}
\loss_{DG}(t,\mathbf{d},\Theta) = \lambda_0\loss_0(t,\mathbf{d},\Theta) 
+ \lambda_1 \reg_1(\jacobian_{\loss_1}(t,\mathbf{d},\Theta))
+ \cdots 
+ \lambda_m \reg_m(\jacobian_{\loss_m}(t,\mathbf{d},\Theta))
\end{align*}
\\
\\
\noindent
where $\mathbf{d} = (d_1,d_2,...d_k)$ is a data sample, $t$ is its corresponding label/target and $\Theta= \{W_1,W_2,...,W_K\}$ represents the parameters of a $K$ layer neural network.\footnote{The loss $\loss_0$ could be a single objective (as in minimizing negative log likelihood for a classification task) or an objective combined with an auxiliary task, as in multi-task learning. In the latter setting, $\reg_1$ could be the regularizer for the main task and $\reg_2$ could be an optional regularizer on the secondary task, especially if the model needs to be robust with respect to that task as well.} We use  $\jacobian_{\loss_i}$ to denote the gradient of $\loss_i$ (the gradient of $\loss_i$ with respect to $\mathbf{d}$). $\lambda_0,\lambda_1,\dots,\lambda_m$ are the weight coefficients of the terms in the DataGrad loss function. \black{Close to our work, \citep{lyuunify} present a heuristic way to optimize a special case of this objective. By directly provding an algorithm, our analysis can explain what their algorithm optimizes.}
%
%
%

We denote the entire dataset as $\mathcal{D}=\{(\mathbf{d}^{(1)}, t^{(1)}),\dots, (\mathbf{d}^{(n)}, t^{(n)})\}$. Following the framework of empirical risk minimization with stochastic gradient descent, the goal is to minimize the objective function: $\sum\limits_{i=1}^n \loss_{DG}(t^{(i)},\mathbf{d}^{(i)},\Theta)$
by iterating the following parameter updates (here $w^{\ell}_{ij}$ is the component of $\Theta$ representing the weight of the incoming edge to node $i$ of layer $\ell$ from node $j$ of layer $\ell-1$):
\begin{align}
w^{\ell}_{ij} \leftarrow w^\ell_{ij} - \eta \lambda_0\frac{\partial}{\partial w^{(\ell)}_{ij}}\left[\loss_0(t,\mathbf{d},\Theta)\right] - \eta\sum\limits_{r=1}^m \lambda_r \frac{\partial}{\partial w^{(\ell)}_{ij}}\left[\reg_r(\jacobian_{\loss_r}(t,\mathbf{d},\Theta)\right] \label{eqn:updateeqn}
\end{align}
\noindent
where $\eta$ is the step-size coefficient.
\subsection{The Derivation}\label{sec:derivation}
 The first update term of Equation \ref{eqn:updateeqn}, $\frac{\partial}{\partial w^{(\ell)}_{ij}}\left[\loss_0(t,\mathbf{d},\Theta)\right]$, is provided by standard back-propagation. For the remaining terms, since the gradient of the loss also depends on the current weights $\Theta$, we see that
\begin{align}
\frac{\partial \reg_r(\jacobian_{\loss_r}(t,\mathbf{d},\Theta))}{\partial w^{(\ell)}_{ij}}
 &= \frac{\partial \reg_r(\frac{\partial \loss_r}{\partial d_1},\dots,\frac{\partial \loss_r}{\partial d_k})}{\partial w^{(\ell)}_{ij}}
=\sum\limits_{s=1}^k \frac{\partial \reg_r(a_1,\dots,a_k)}{\partial a_s} \frac{\partial^2 \loss_r(t,\mathbf{d},\Theta)}{\partial  w^{(\ell)}_{ij}\partial d_s}\label{eqn:partialderiva}
\end{align}
where $a_s$ is a variable that takes the current value of $\frac{\partial \loss_r}{\partial d_k}$.
It turns out that these mixed partial derivatives (with respect to weights and with respect to data) have structural similarities to the Hessian (since derivatives with respect to the data are computed almost exactly the same way as the derivatives with respect to the lowest layer weights). Since exact computation of the Hessian is slow \citep{bishophessian}, we would expect that the computation of this matrix of partial derivatives would also be slow. However, it turns out that we do not need to compute the full matrix -- we only need this matrix times a vector, and hence we can use ideas reminiscent of fast Hessian multiplication algorithms \citep{multiplyhessian}. At points of continuous differentiability, we have:
\begin{align}
\sum\limits_{s=1}^k \frac{\partial \reg_r(a_1,\dots,a_k)}{\partial a_s} \frac{\partial^2 \loss_r(t,\mathbf{d},\Theta)}{\partial  w^{(\ell)}_{ij}\partial d_s}&=\sum\limits_{s=1}^k \frac{\partial \reg_r(a_1,\dots,a_k)}{\partial a_s} \frac{\partial^2 \loss_r(t,\mathbf{d},\Theta)}{\partial d_s\partial  w^{(\ell)}_{ij}}\nonumber\\
&= \frac{\partial^2 \loss(t, \mathbf{d}+\phi\mathbf{y},\Theta)}{\partial \phi~\partial w^\ell_{ij}} \label{eqn:partialderivb}
\end{align}
evaluated at the point $\phi=0$ and direction $\mathbf{y}=( \frac{\partial \reg_r(a_1,\dots,a_k)}{\partial a_1}, \dots,  \frac{\partial \reg_r(a_1,\dots,a_k)}{\partial a_k})$.\footnote{Note that Equation \ref{eqn:partialderivb} follows from the chain rule.} The outer directional derivative with respect to the scalar $\phi$ can be computed using finite differences.
Thus, Equations \ref{eqn:partialderiva} and \ref{eqn:partialderivb} mean that we can compute the term $\frac{\partial}{\partial w^{(\ell)}_{ij}}\left[\reg_r(\jacobian_{\loss_r}(t,\mathbf{d},\Theta)\right]$ from the stochastic gradient descent update equation (Equation \ref{eqn:updateeqn}) as follows.

\begin{enumerate}
\item Use standard back-propagation to simultaneously compute the vector derivatives $\frac{\partial \loss_r(t,\mathbf{d},\Theta)}{\partial \Theta}$ and  $\frac{\partial \loss_r(t,\mathbf{d},\Theta)}{\partial \mathbf{d}}$ (note that the latter corresponds to the vector $(a_1,\dots,a_s)$ in our derivation).
\item Analytically determine the gradient of $\reg_r$ with respect to its immediate inputs. For example, if $\reg_r$ is the $L_2$ penalty $\reg_r(x_1,\dots,x_s)=|x_1|^2+\cdots+|x_s|^2$ then the immediate gradient would be $(2x_1, \dots, 2x_s)$ and if $\reg_r$ is the $L_1$ penalty, the immediate gradient would be $(\text{sign}(x_1),\dots,\text{sign}(x_s))$.
\item Evaluate the immediate gradient of $\reg_r$ at the vector  $\frac{\partial \loss_r(t,\mathbf{d},\Theta)}{\partial \mathbf{d}}$. This corresponds to the adversarial direction, as is denoted by $\mathbf{y}$ in our derivation.
\item Form the adversarial example $\widehat{\mathbf{d}} = \mathbf{d}+\phi\mathbf{y}$, where $\mathbf{y}$ is the result of the previous step and $\phi$ is a small constant.
\item Use a second back-propagation pass (with $\widehat{\mathbf{d}}$ as input) to compute  $\frac{\partial \loss_r(t,\widehat{\mathbf{d}},\Theta)}{\partial \Theta}$ and then return the finite difference $\left( \frac{\partial \loss_r(t,\widehat{\mathbf{d}},\Theta)}{\partial \Theta}  -  \frac{\partial \loss_r(t,\mathbf{d},\Theta)}{\partial \Theta}\right)/\phi$
\end{enumerate}

\subsection{The High-level View: Putting it All Together}\label{sec:together}
At a high level, the loss $\loss_r$ and regularizer $\reg_r$ together serve to define an adversarial noise vector $\mathbf{y}$ and adversarial example $\widehat{\mathbf{d}}=\mathbf{d}+\phi \mathbf{y}$ (where $\phi$ is a small constant), as explained in the previous section. Different choices of $\loss_r$ and $\reg_r$ result in different types of adversarial examples. For example, setting $\reg_r$ to be the $L_1$ penalty, the resulting adversarial example is the same as the \emph{fast gradient sign method} of \cite{goodfellow_explaining_2014}.

Putting together the components of our finite differences algorithm, the stochastic gradient descent update equation becomes:
\begin{align}
w^{\ell}_{ij} &\leftarrow w^\ell_{ij} - \eta \lambda_0\frac{\partial \loss_0(t,\mathbf{d},\Theta)}{\partial w^{(\ell)}_{ij}} - \eta\sum\limits_{r=1}^m \frac{\lambda_r}{\phi}\left(\frac{\partial \loss_r(t,\mathbf{x},\Theta)}{\partial w^{(\ell)}_{ij}}\Big|_{\mathbf{x}=\mathbf{x}_r}-\frac{\partial \loss_r(t,\mathbf{d},\Theta)}{\partial w^{(\ell)}_{ij}}\right)  \nonumber\\
&= w^\ell_{ij} - \eta \left(\lambda_0 - \sum_r\frac{\lambda_r}{\phi}\right)\frac{\partial \loss_0(t,\mathbf{d},\Theta)}{\partial w^{(\ell)}_{ij}} - \eta\sum\limits_{r=1}^m \frac{\lambda_r}{\phi}\frac{\partial \loss_r(t,\mathbf{x},\Theta)}{\partial w^{(\ell)}_{ij}}\Big|_{\mathbf{x}=\mathbf{x}_r}
\label{eqn:updateeqn2}
\end{align}
where $\mathbf{x}_r$ is the adversarial example of $\mathbf{d}$ resulting from regularizer $\reg_r$ in conjunction with loss $\loss_r$, and the notation $\frac{\partial \loss_r(t,\mathbf{x},\Theta)}{\partial w^{(\ell)}_{ij}}\Big|_{\mathbf{x}=\mathbf{x}_r}$ here specifically means to compute the derivative using back-propagation with $\mathbf{x}_r$ as an input -- in other words, $\mathbf{x}_r$ is not to be treated as a function of $\Theta$ (and its components  $w^{(\ell)}_{ij}$ ) when computing this partial derivative.

\subsection{How Prior Works Are Instances of Datagrad}
\label{related_work}
Since the recent discovery of adversarial samples \citep{szegedy_intriguing_2013}, a variety of remedies have been proposed to make neural architectures robust to this problem.  A straightforward solution is to simply add adversarial examples during each training round of stochastic gradient descent  \citep{szegedy_intriguing_2013}. This is exactly what Equation \ref{eqn:updateeqn2} specifies, so that post-hoc  solution can be justified as a regularization of the data gradient. Subsequent work \citep{goodfellow_explaining_2014} introduced the objective function
 $\sum_d \alpha\loss(t,\mathbf{d},\Theta) + (1-\alpha)\loss(t,\widehat{\mathbf{d}},\Theta)$, where $\widehat{\mathbf{d}}$ is the adversarial version of input $d$.  A gradient-based method would need to compute the derivative with respect to $w^{(\ell)}_{ij}$, which is $\alpha\frac{\partial \loss(t,d,\Theta)}{\partial w^{(\ell)}_{ij}} + (1-\alpha)\frac{\partial \loss(t,\widehat{\mathbf{d}},\Theta)}{\partial w^{(\ell)}_{ij}} + (1-\alpha)\frac{\partial \loss(t,\widehat{d},\Theta)}{\partial \widehat{\mathbf{d}}} \cdot \frac{d~ \widehat{\mathbf{d}}}{d w^{(\ell)}_{ij}}$, since the construction of $\widehat{\mathbf{d}}$ depends on $w^{(\ell)}_{ij}$.  Their work approximates the optimization by ignoring the third term, as it is difficult to compute.  This approximation then results in an updated equation having the form of Equation \ref{eqn:updateeqn2}, and hence actually optimizes the DataGrad objective.
\cite{nokland_improving_2015} present a variant where the deep network is trained using back-propagation only on adversarial examples (rather than a mix of adversarial and original examples). Equation 4 shows that this method optimizes the DataGrad objective with \black{$r=1$ and $\lambda_0$ and $\lambda_1$ chosen so that the $\frac{\partial \loss_0(t,\mathbf{d},\Theta)}{\partial w^{(\ell)}_{ij}}$ term is eliminated.}

Both  \cite{huang_learning_2015} and  \cite{miyato_distributional_2015} propose to optimize constrained objective functions that can be put in the form $\min_{\Theta}\sum_{\mathbf{d}} \max_{g(r)\leq c} f(t,\mathbf{d},r,\Theta)$, where $r$ represents adversarial noise and the constraint $g(r)\leq c$ puts a bound on the size of the noise. Letting $r^*(\mathbf{d},\Theta)$ be the (constrained) optimal value of $r$ for each $\mathbf{d}$ and setting of $\Theta$, this is the same as the objective $\min_{\Theta}\sum_{\mathbf{d}}f(t,\mathbf{d},r^*(\mathbf{d},\Theta),\Theta)$. The derivative of any term in the summation respect to $w^{(\ell)}_{ij}$ is then equal to 
\begin{align}
\frac{\partial f(t,\mathbf{d},r,\Theta)}{\partial w^{(\ell)}_{ij}}\Big|_{r=r^*(\mathbf{d},\Theta)} + \frac{\partial f(t,\mathbf{d},r,\Theta)}{\partial r}\Big|_{r=r^*(\mathbf{d},\Theta)}\cdot \frac{\partial r^*(\mathbf{d,\Theta})}{\partial w^{(\ell)}_{ij}}\label{eqn:constrainedopt}
\end{align}
Now, if $r^*(\mathbf{d},\Theta)$ were an unconstrained maximum value of $r$, then  $\frac{\partial f(t,\mathbf{d},r,\Theta)}{\partial r}\Big|_{r=r^*(\mathbf{d},\Theta)}$ would equal $0$ and the second term of Equation \ref{eqn:constrainedopt} would disappear. However, since $r^*$ is a constrained optimum and the constraint is active, the second term would generally be nonzero. Since the derivative of the constrained optimum is difficult to compute, \cite{huang_learning_2015} and  \cite{miyato_distributional_2015} opt to approximate/simplify the derivative making the second term disappear (as it would in the unconstrained case). \black{Comparing the remaining term to Equation \ref{eqn:updateeqn2} shows that they are optimizing the DataGrad objective with
$r=1$ and $\lambda_0$ and $\lambda_1$ carefully chosen to eliminate the $\frac{\partial \loss_0(t,\mathbf{d},\Theta)}{\partial w^{(\ell)}_{ij}}$ term.}

\black{In an approach that  ends up closely related to ours, \citep{lyuunify} consider the objective $\min_{\theta}\max_{r:||r||_p\leq\sigma} \loss(x+r;\theta)$ and a linearized inner version $\max_{r:||r||_p\leq\sigma} \loss(x)+\grad_x\loss^T r$. They iteratively select $r$ by optimizing the latter and $\theta$ by back-propagation on the former (with $r$ fixed). Since the $\theta$ update is not directly minimizing the linearized objective, \citep{lyuunify} claimed the procedure was only an approximation of what we call the DataGrad objective. However, their method devolves to training on adversarial examples, so as before, Equation \ref{eqn:updateeqn2} shows they are actually optimizing the DataGrad objective but with
$r=1$ and $\lambda_0$ and $\lambda_1$ carefully chosen to eliminate the $\frac{\partial \loss_0(t,\mathbf{d},\Theta)}{\partial w^{(\ell)}_{ij}}$ term.}



Finally, \cite{gu_towards_2014}  penalizes the Frobenius norm of the deep gradient. However, they do this with a shallow layer-wise approximation. Specifically, they note that shallow contractive auto-encoders optimize the same objective for shallow (1-layer) networks and that the gradient of the gradient can be computed analytically in those cases \cite{gu_towards_2014}. Thus,  \cite{gu_towards_2014} applies this penalty layer by layer (hence it is a penalty on the derivative of each layer with respect to its immediate inputs) and uses this penalty as an approximation to regularizing the deep gradient. Since \emph{DataGrad} does regularize the deep gradient, the work of \cite{gu_towards_2014} can also be viewed as an approximation to \emph{DataGrad}.

Thus, DataGrad provides a unifying view of previously proposed optimizations for training deep architectures that are resilient to adversarial noise.

%


\begin{figure}
\begin{subfigure}{.5\textwidth}
  \centering
  \includegraphics[width=\linewidth]{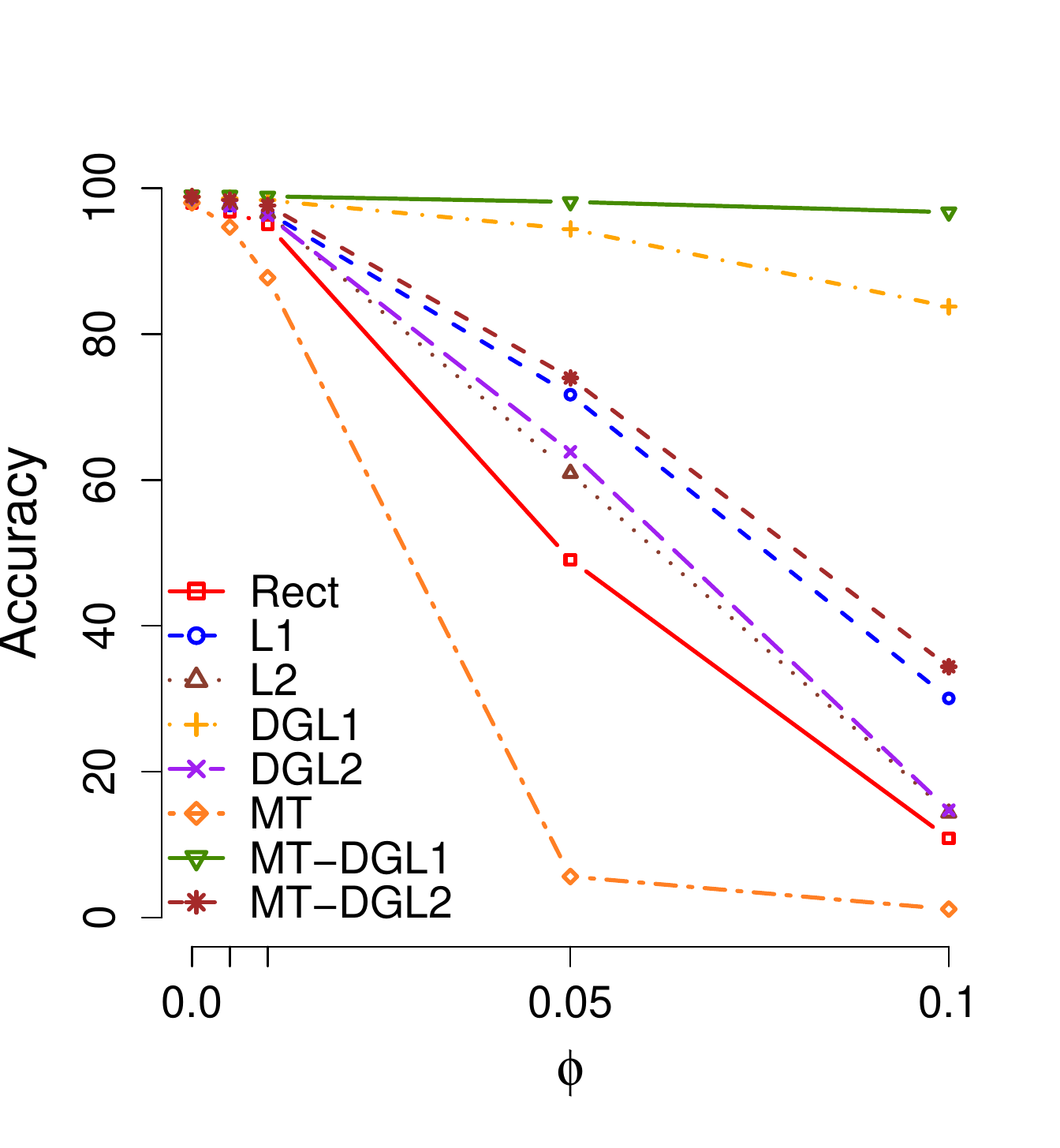}
  \caption{Performance on L1 adversarial samples.}
  \label{l1_noise}
\end{subfigure}%
\begin{subfigure}{.5\textwidth}
  \centering
  \includegraphics[width=\linewidth]{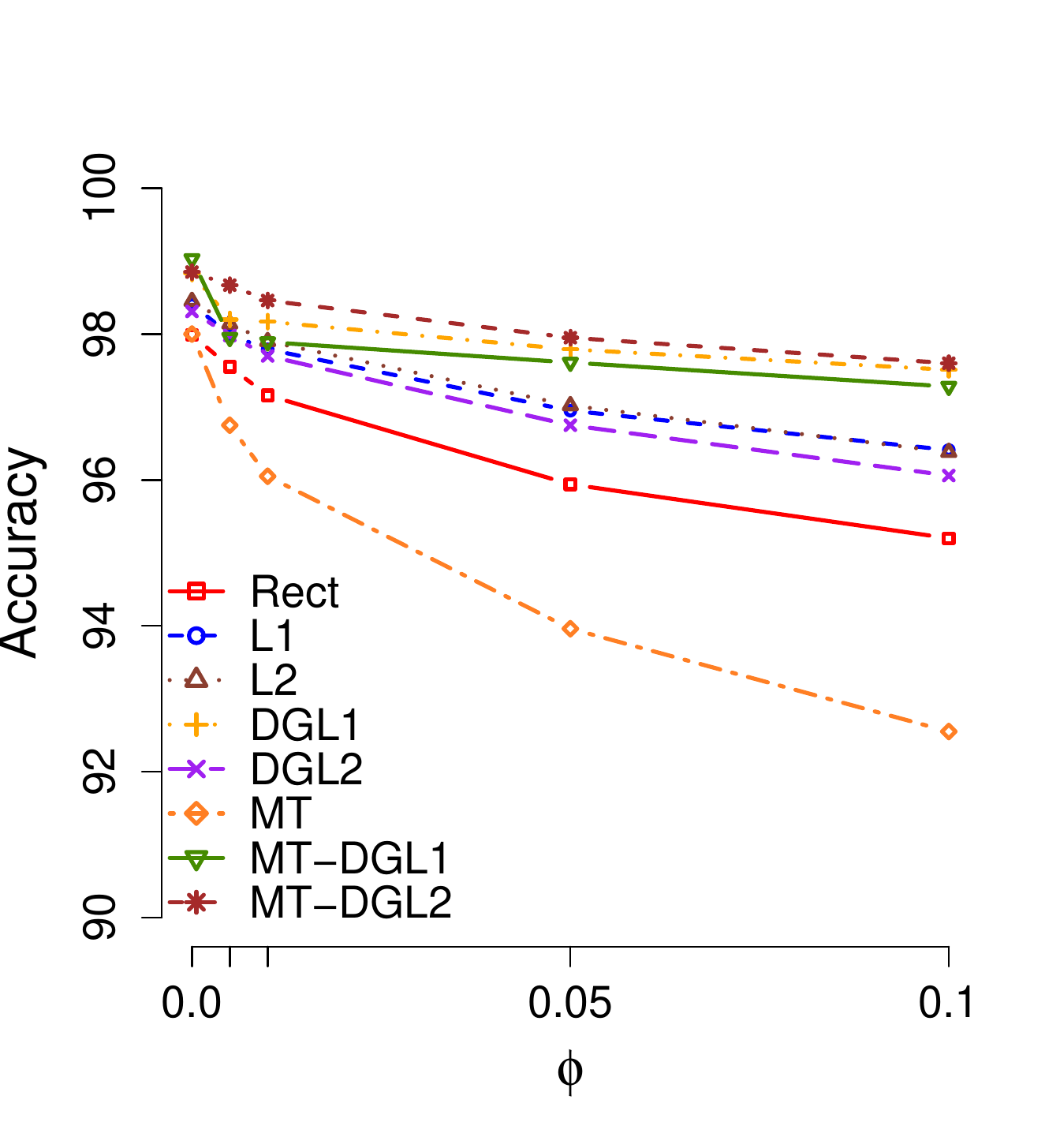}
  \caption{Performance on L2 adversarial samples.}
  \label{l2_noise}
\end{subfigure}
\caption{Model performance when each model is its own adversary.  Note that $\phi$ on the x-axis indicates degree of ($\reg_1$ or $\reg_2$) noise used to create adversarial samples. Terms in the legend refer to specific architectures (e.g., \emph{L2} refers to the L2-regularized network).}
\label{adv_error_curves}
\end{figure}

\section{Experimental Results}
\label{results}
Given that we have shown that previous approaches are instances of the general DataGrad framework, it is not our intention to replicate prior work. Rather, we intend to not only test the effectiveness of our finite difference approximation but to also show that one can flexibly use DataGrad in other scenarios, such as adding multi-task cues within the adversarial framework. To test the proposed DataGrad framework, we conduct experiments using the permutation-invariant MNIST data-set~\footnote{http://yann.lecun.com/exdb/mnist/.}, comprised of 60,000 training samples and 10,000 testing samples. A validation subset of 10,000 samples (randomly sampled without replacement from the training split) was used for tuning architecture meta-parameters via a coarse grid-search. Image features were gray-scale pixel values of which we normalized to the range of $[0,1]$. We find that turning our attention first to an image classification problem like MNIST is appropriate since the adversarial problem was first presented in the context of computer vision problems. Investigation of our framework's usefulness in domains such as text is left for future work.

In this study, we experiment with two concrete instantiations of the DataGrad framework, namely DataGrad-L1 (\emph{DGL1}) and DataGrad-L2 (\emph{DGL2}). By setting $\lambda_0=1$, letting $\lambda_1$ freely vary as a meta-parameter and $\lambda_j=0$ for $j>1$, and $\loss_0=\loss_1$, choosing $\reg_1$ to be the L1 penalty results in \emph{DGL1} while choosing the L2 penalty yields \emph{DGL2}. As a result, DataGrad becomes a regularization algorithm on either the $L_1$ or $L_2$ norm of the gradient of the loss $\loss_0$. In this setup, DataGrad requires two forward passes and two backward passes to perform a weight update. 

We are interested in evaluating how DataGrad compares to conventional and non-conventional forms of regularization. Beyond traditional $L_1$ and $L_2$ regularization of the network parameters ($L1$ and $L2$, respectively), we also experimented with the regularizing effect that multi-task learning (\emph{M T}) has on parameter learning in the interest of testing whether or not having an auxiliary objective could introduce any robustness to adversarial samples in additional to improved generalization. In order to do so, we designed a dual-task rectifier network with two disjoint sets of output units, each connected to the penultimate hidden layer by a separate set of parameters (i.e, $U_0$ for task 0, $U_1$ for task 1). The leftmost branch is trained to predict one of the 10 original target digit labels associated with an image (as in the original MNIST task set-up) which corresponds to loss $\loss_0$, while the rightmost branch is trained to predict one of five artificially constructed categories pertaining to the discretized degree of rotation of the image, which corresponds to loss $\loss_1$.\footnote{This auxiliary task consists of essentially predicting whether a given sample has been artificially rotated $0^{\circ}$, $15^{\circ}$ to the left, $30^{\circ}$ to the left, $15^{\circ}$ to the right, or $30^{\circ}$ to the right. To automatically generate these auxiliary labels, we artificially rotate each sample in the training set 5 times (once for each category) and record that rotation as a label. This could be viewed as a form of data-set expansion, which generally leads to improved generalization ability. However, since we assign an artificial label from a different task to accompany each sample, we use the expanded training set to create a different optimization problem.} The multi-objective optimization problem for this set-up then becomes:

\begin{equation}
\label{mt_loss}
\loss_{DG}(t,\mathbf{d},\Theta) = \loss_0(t,\mathbf{d},\Theta) + \gamma\loss_1(t,\mathbf{d},\Theta) 
+ \lambda_1 \reg_1(\jacobian_{\loss_0}(t,\mathbf{d},\Theta))
\end{equation}

\noindent
where $\gamma$ is a coefficient that controls the influence of the auxiliary objective $\loss_1$ on the overall parameter optimization problem. Note that we have extended Equation \ref{mt_loss} to include a DataGrad term, which may either be of L1 form, \emph{MT-DGL1}, or of L2 form, \emph{MT-DGL2}. All regularized architectures are compared against the baseline sparse rectifier network, \emph{Rect}.

\begin{figure}
\begin{subfigure}{.5\textwidth}
  \centering
  \includegraphics[width=\linewidth]{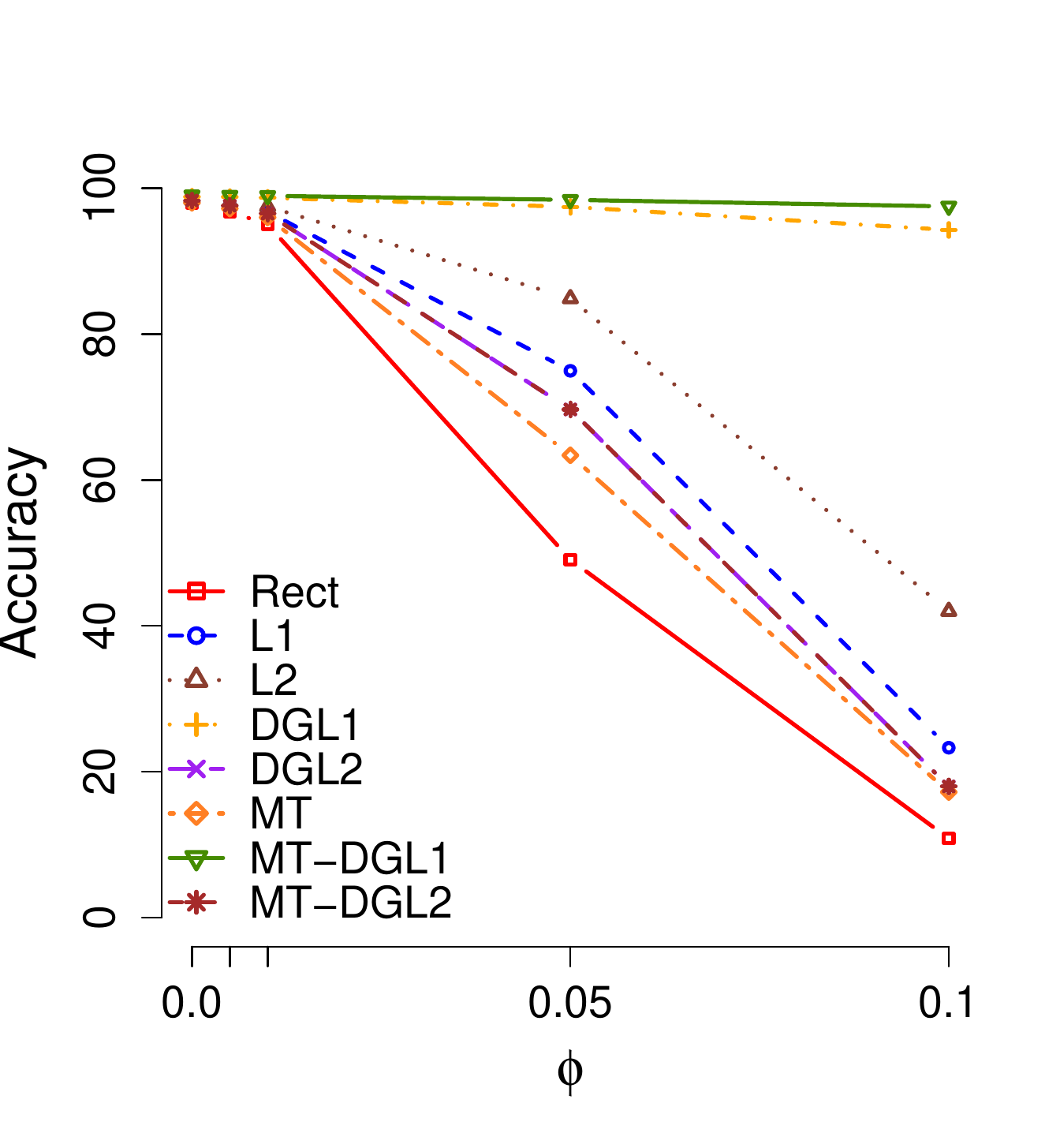}
  \caption{Simple rectifier network is the adversary.}
  \label{rect_attack}
\end{subfigure}%
\begin{subfigure}{.5\textwidth}
  \centering
  \includegraphics[width=\linewidth]{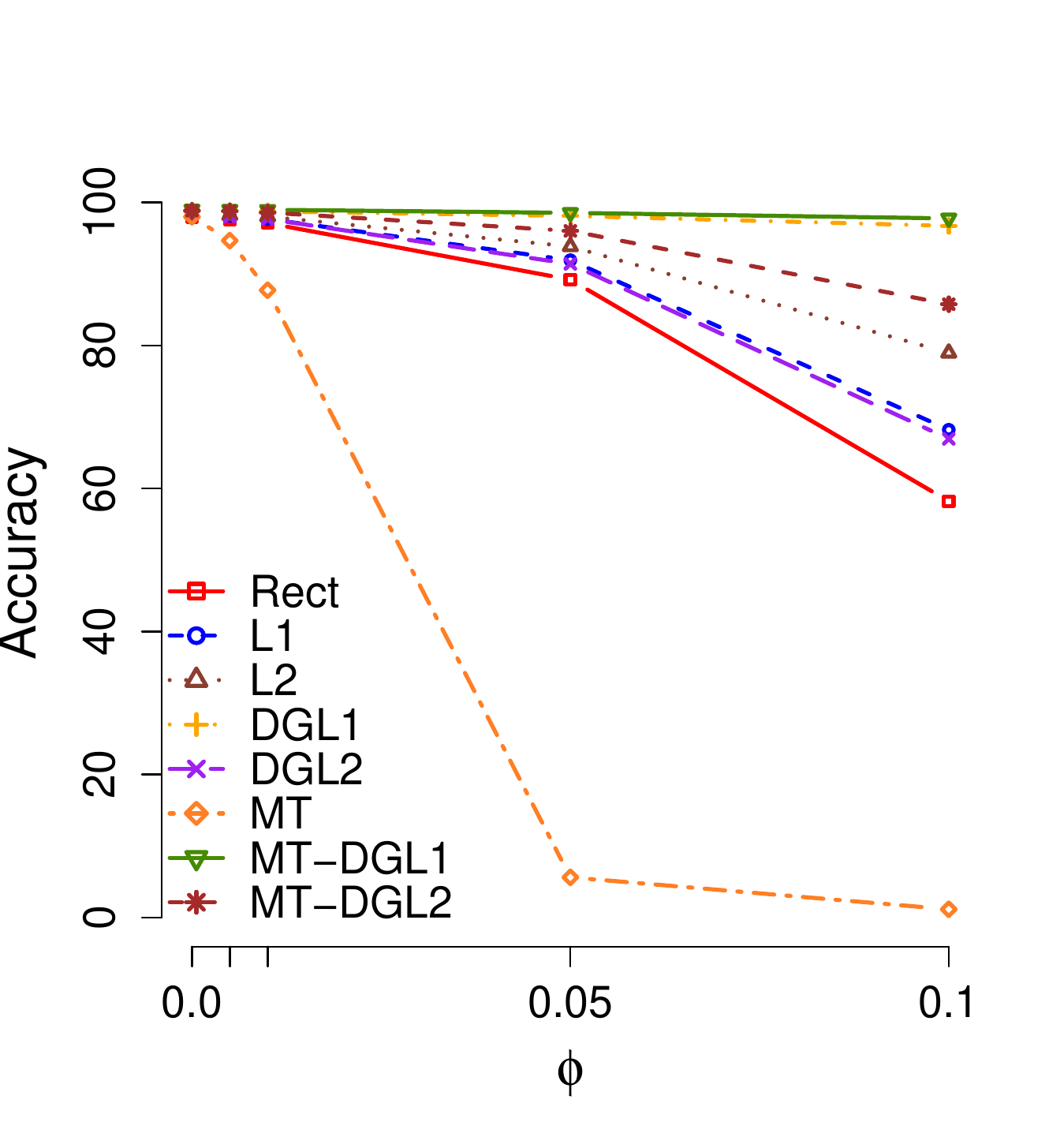}
  \caption{Multi-task rectifier network is the adversary.}
  \label{mt_rect_attack}
\end{subfigure}
\caption{Model performance when the adversarial architecture is the same, either simple (\emph{Rect}) or multi-task (\emph{MT}).  Note that $\phi$ on the x-axis indicates degree of $\reg_1$ noise applied to create adversarial samples.}
\label{adv_error_curves_2}
\end{figure}

We implemented several deep sparse rectifier architectures \citep{glorot_deep_2011} (3 hidden layers, each with 784 latent variables--parameters were initialized following the scheme of \cite{he2015delving}), which were all to be trained in a gradient descent framework under the various regularization schemes described earlier. Mini-batches of size 100 were used for calculating each parameter update. Hyper-parameters and ranges searched included  the $\lambda_1 = [0.0001, 0.1]$ and $\phi = [0.005, 0.1]$ coefficients for controlling DataGrad, the $L1 = [0.0001, 0.01]$ and $L2 = [0.0001, 0.01]$ penalty coefficients (which would simply appear as $\lambda$, as in Appendix A) for controlling the classical regularization terms, the $\gamma = [0.25,0.75]$ auxiliary objective weight, and the gradient descent step-size $\alpha = [0.001,0.2]$. We did not use any additional gradient-descent heuristics (i.e., momentum, adaptive learning rates, drop-out, etc.) for simplicity, since we are interested in investigating the effect that the regularizers have on model robustness to adversarial samples.

To evaluate the above architectures in the adversarial setting, we conduct a series of experiments where each trained model plays the role of ``attacker''. An adversarial test-set, comprised of 10,000 samples, is generated from the attacking model via back-propagation, using the derivative of the loss with respect to the model's inputs followed by the application of the appropriate regularizer function (either $\reg_1$ or $\reg_2$) to create the noise. The amount of noise applied is controlled by $\phi$, which we varied along the values $\{0.0, 0.005, 0.01, 0.05, 0.1\}$, which corresponds to maximal pixel gains of $\{0, \sim 1, \sim 3, \sim 12, \sim 25\}$ (0 would mean be equivalent to using the original test-set). Generalization performances reported in all figures in this section are of the architectures that achieved best performance on the validation subset (consult Appendix A for a full treatment of performance across a range of $\phi$ and $\lambda$ values). 

\begin{figure}
\centering
\begin{subtable}{\textwidth}
  \centering
  \resizebox{0.7\hsize}{!} {%
  \begin{tabular}{lrrrrr}
  \multicolumn{1}{l}{\textbf{$\phi$ =}}&\multicolumn{1}{c}{\begin{tabular}[x]{@{}c@{}}\textbf{0.0} \\\end{tabular}}&\multicolumn{1}{c}{\begin{tabular}[x]{@{}c@{}}\textbf{0.005}\\\end{tabular}}&\multicolumn{1}{c}{\begin{tabular}[x]{@{}c@{}}\textbf{0.01}\\\end{tabular}}&\multicolumn{1}{c}{\begin{tabular}[x]{@{}c@{}}\textbf{0.05}\\\end{tabular}}&\multicolumn{1}{c}{\begin{tabular}[x]{@{}c@{}}\textbf{0.1}\\\end{tabular}}\tabularnewline
  \hline
  \textit{Rect-MLP} & $97.99 \%$ & $96.80 \%$ & $95.01 \%$ & $49.06 \%$ & $10.83 \%$\tabularnewline
  \hline
  \end{tabular}
  }
  \label{adv_norm_error}
\end{subtable}%
\newline
\newline
\begin{subtable}{\textwidth}
  \centering
  \begin{tabular}{@{\hskip 1.75cm}c@{\hskip 0.75cm}c@{\hskip 0.75cm}c@{\hskip 0.75cm}c@{\hskip 0.75cm}c}
  \includegraphics[scale=0.95]{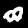}&\includegraphics[scale=0.95]{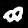}&\includegraphics[scale=0.95]{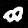}&\includegraphics[scale=0.95]{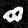}&\includegraphics[scale=0.95]{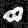}\\
  \end{tabular}
  \label{adv_norm_samps}
\end{subtable}

  \begin{subtable}{\textwidth}
  \centering
  \resizebox{.7\hsize}{!} {%
  \begin{tabular}{lrrrrr}
  \multicolumn{1}{l}{\textbf{$\phi$ =}}&\multicolumn{1}{c}{\begin{tabular}[x]{@{}c@{}}\textbf{0.0} \\\end{tabular}}&\multicolumn{1}{c}{\begin{tabular}[x]{@{}c@{}}\textbf{0.005}\\\end{tabular}}&\multicolumn{1}{c}{\begin{tabular}[x]{@{}c@{}}\textbf{0.01}\\\end{tabular}}&\multicolumn{1}{c}{\begin{tabular}[x]{@{}c@{}}\textbf{0.05}\\\end{tabular}}&\multicolumn{1}{c}{\begin{tabular}[x]{@{}c@{}}\textbf{0.1}\\\end{tabular}}\tabularnewline
  \hline  
  \textit{Rect-L1} & $98.41 \%$ & $97.62 \%$ & $96.68 \%$ & $71.69 \%$ & $30.05 \%$\tabularnewline
  \hline
  \end{tabular}
  }
  \label{adv_l2_error}
\end{subtable}%
\newline
\newline
\begin{subtable}{\textwidth}
  \centering
  \begin{tabular}{@{\hskip 1.75cm}c@{\hskip 0.75cm}c@{\hskip 0.75cm}c@{\hskip 0.75cm}c@{\hskip 0.75cm}c}
  \includegraphics[scale=0.95]{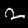}&\includegraphics[scale=0.95]{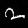}&\includegraphics[scale=0.95]{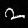}&\includegraphics[scale=0.95]{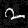}&\includegraphics[scale=0.95]{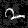}\\
  \end{tabular}
  \label{adv_l2_samps}
\end{subtable}

  \begin{subtable}{\textwidth}
  \centering
  \resizebox{.7\hsize}{!} {%
  \begin{tabular}{lrrrrr}
  \multicolumn{1}{l}{\textbf{$\phi$ =}}&\multicolumn{1}{c}{\begin{tabular}[x]{@{}c@{}}\textbf{0.0} \\\end{tabular}}&\multicolumn{1}{c}{\begin{tabular}[x]{@{}c@{}}\textbf{0.005}\\\end{tabular}}&\multicolumn{1}{c}{\begin{tabular}[x]{@{}c@{}}\textbf{0.01}\\\end{tabular}}&\multicolumn{1}{c}{\begin{tabular}[x]{@{}c@{}}\textbf{0.05}\\\end{tabular}}&\multicolumn{1}{c}{\begin{tabular}[x]{@{}c@{}}\textbf{0.1}\\\end{tabular}}\tabularnewline
  \hline
  \textit{Rect-DG} & $98.83 \%$ & $98.63 \%$ & $98.34 \%$ & $94.41 \%$ & $83.74 \%$\tabularnewline
  \hline
  \end{tabular}
  }
  \label{adv_datagrad_error}
\end{subtable}%
\newline
\newline
\begin{subtable}{\textwidth}
  \centering
  \begin{tabular}{@{\hskip 1.75cm}c@{\hskip 0.75cm}c@{\hskip 0.75cm}c@{\hskip 0.75cm}c@{\hskip 0.75cm}c}
  \includegraphics[scale=0.95]{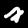}&\includegraphics[scale=0.95]{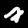}&\includegraphics[scale=0.95]{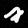}&\includegraphics[scale=0.95]{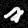}&\includegraphics[scale=0.95]{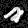}\\
  \end{tabular}
  \label{adv_datagrad_samps}
\end{subtable}

\caption{Adversarial test-set accuracy ($\phi = 0$ corresponds to original test-split) and samples generated from a deep sparse rectifier network in the case of (starting from top of diagram to bottom):  (1), no regularization, (2), L1-regularization, and (3), L1 DataGrad-regularization. The measures reported here are when each model is used to attack itself (akin to the malicious user using the exact same architecture to generate samples).}
\label{adv_viz_results}
\end{figure}


We observe in Figures \ref{adv_error_curves}, \ref{adv_error_curves_2}, and \ref{adv_viz_results} that a DataGrad-regularized architecture outperforms the non-regularized baseline as well as alternatively-regularized ones. Note that the accuracy in Figure \ref{adv_error_curves} only drops to as low as 92\% in the worst case, meaning that $\reg_2$ samples seem to cause only minimal damage and should be much less of a concern than $\reg_1$ samples (which would be akin to generating noise via the fast gradient sign method).\footnote{Note that in the case of $\reg_2$ noise, DataGrad-L2 yields slightly more robust performance in this scenario, closely followed by DataGrad-L1, which makes intuitive sense.} With respect to using only an auxiliary objective to regularize the model (\emph{MT}), in both Figures \ref{adv_error_curves}, \ref{adv_error_curves_2}, we often see that a dual-task model performs the worst when adversarial noise is introduced, surprisingly even more so than the simple rectifier baseline. Figure \ref{adv_error_curves_2} shows that when the non-regularized multi-task architecture is attacked by itself, its error can drop as low as nearly 1\%. However, when a DataGrad term is added to the multi-task objective, we achieve nearly \emph{no} loss in classification performance. This means that a multi-task, DataGrad-regularized rectifier network appears to be quite robust to adversarial samples (of either $\reg_1$ or $\reg_2$ form) either generated from itself or other perceptron architectures (\emph{DGL1} moreso than \emph{DGL2}).

Classical $L_1$ and $L_2$ regularizers appear to to mitigate some of the damage in some instances, but seemingly only afford at best only modest robustness to adversarial perturbation.  In contrast, the proposed \emph{DGL1} and \emph{DGL2} regularizers appear to yield a significant reduction in error on all of the various adversarial test-sets, the improvement clearer as $\phi$ is increased (as evidenced in Figure \ref{adv_viz_results}).  The visualization of some adversarial samples in Figure \ref{adv_viz_results} demonstrates that even when more noise is applied to generate stronger adversarials, the samples themselves are still quite recognizable to the human eye.  However, a neural architecture, such as a deep rectifier network, is sensitive to adversarial noise and incorrectly classifies these images. In addition to robustness against adversarial samples, we also observe improved classification error on the original test-set when using \emph{DataGrad} or multi-task \emph{DataGrad}, the \emph{DGL1} and \emph{MT-DGL1} variants offering the lowest error of all. For further experimental results exploring the performance and sensitivity of DataGrad to its meta-parameters (including when other architectures are the adversary), we defer the reader to Appendix A.

\section*{Conclusion}
Here we have shown how previous proposals can be viewed as instances of a simple, general framework and provide an efficient, deterministic adversarial training procedure, \emph{DataGrad}. The simplicity of the framework allows for easy extensions, such as adding multi-task cues as another signal to be combined with adversarial training. Empirically, we found that general DataGrad regularization not only significantly reduces error (especially so when combined with a multi-task learning objective) in classifying adversarial samples but also improves generalization.  We postulate a reason for this is that adversarial samples generated during the \emph{DataGrad} learning phase potentially cover more of the underlying data manifold (yielding benefits similar to data-set expansion). 

Since \emph{DataGrad} is effectively a ``deep'' data-driven penalty, it may be used in tandem with most training objective functions (whether supervised, unsupervised \cite{bengio_greedy_2007}, or hybrid \cite{ororbia_deep_hybrid_2015a}). Future work entails further improving the efficiency of the proposed \emph{DataGrad} back-propagation procedure and investigating our procedure in a wider variety of settings.




\bibliographystyle{apa}
\bibliography{ref}

\newpage
\section*{Appendix A: Detailed Results}
\label{append_a}

In this appendix, to augment the experimental results presented in Section \ref{results}, we present the generalization performances of the regularized (and non-regularized models) under various settings of their key hyper-parameters. This particularly applies to $\lambda$ and $\phi$ (when applicable). All model performances reported are those with the learning rate $\alpha$ and auxiliary objective weight $\gamma$ meta-parameters fixed at the value that yielded best validation set performance. Furthermore, each table represents a different adversarial scenario, where a different architecture was selected to be the generator of adversarial examples. In each table, two lines are bolded, one for the single-task architecture and one for the multi-task model that achieves the most robust performance across all noise values of $\reg_1$. We do not report the same adversarial scenarios for $\reg_2$ noise, as we found that it had little impact on model generalization ability (as was seen in Section \ref{results}).

One key observation to take from this set of experiments on the MNIST dataset is that DataGrad, particularly the L1 form, achieves the greatest level of robustness to adversarial samples in all settings when the $\lambda$ and $\phi$ are relatively higher. This is especially so when a DataGrad (L1) term is combined with the multi-task objective.  Note that this appears to be true no matter the adversary (even a DataGrad- or multi-task-regularized one).

We remark that perhaps even further performance improvement could be obtained if one added an additional DataGrad term to the auxiliary objective. In particular, this would apply to the rarer setting when one would also desire additional adversarial robustness with respect to the auxiliary objective. Tables \ref{simple_attack}-\ref{mt_dgl2_attack}, presented over the next several pages, contain the full adversarial setting results.

\begin{table}[!h]
\caption{Comparative results where all models are attacked by a simple rectifier network, \emph{Rect}, using Laplacian ($\reg_1$) adversarial noise.}
\label{simple_attack}
\centering
\resizebox{12.5cm}{!} {%
\begin{tabular}{lrrrrr}
\multicolumn{1}{l}{\begin{tabular}[x]{@{}c@{}}\textbf{Model}\\\end{tabular}}&\multicolumn{1}{c}{\begin{tabular}[x]{@{}c@{}}\textbf{$\phi$ = 0.0}\\\end{tabular}}&\multicolumn{1}{c}{\begin{tabular}[x]{@{}c@{}}\textbf{$\phi$ = 0.005}\\\end{tabular}}&\multicolumn{1}{c}{\begin{tabular}[x]{@{}c@{}}\textbf{$\phi$ = 0.01}\\\end{tabular}}&\multicolumn{1}{c}{\begin{tabular}[x]{@{}c@{}}\textbf{$\phi$ = 0.05}\\\end{tabular}}&\multicolumn{1}{c}{\begin{tabular}[x]{@{}c@{}}\textbf{$\phi$ = 0.1}\\\end{tabular}}\tabularnewline
\hline
\textit{ DGL1 $\lambda =$ 0.0001 $\phi =$ 0.01} & $98.25$ & $97.38$ & $96.14$ & $59.74$ & $14.20$\tabularnewline
\textit{ DGL1 $\lambda =$ 0.0001 $\phi =$ 0.05} & $98.57$ & $98.17$ & $97.76$ & $89.18$ & $59.98$\tabularnewline
\textit{ DGL1 $\lambda =$ 0.0001 $\phi =$ 0.1} & $98.47$ & $98.13$ & $97.75$ & $90.00$ & $70.03$\tabularnewline
\textit{ DGL1 $\lambda =$ 0.001 $\phi =$ 0.01} & $98.03$ & $97.26$ & $96.05$ & $62.77$ & $14.82$\tabularnewline
\textit{ DGL1 $\lambda =$ 0.001 $\phi =$ 0.05} & $98.70$ & $98.43$ & $98.08$ & $92.97$ & $75.65$\tabularnewline
\textit{ DGL1 $\lambda =$ 0.001 $\phi =$ 0.1} & $98.62$ & $98.38$ & $98.11$ & $93.74$ & $83.06$\tabularnewline
\textit{ DGL1 $\lambda =$ 0.01 $\phi =$ 0.01} & $98.36$ & $97.96$ & $97.35$ & $88.03$ & $55.16$\tabularnewline
\textit{ DGL1 $\lambda =$ 0.01 $\phi =$ 0.05} & $98.62$ & $98.50$ & $98.37$ & $95.16$ & $85.78$\tabularnewline
\textit{ DGL1 $\lambda =$ 0.01 $\phi =$ 0.1} & $\mathbf{98.83}$ & $\mathbf{98.77}$ & $\mathbf{98.67}$ & $\mathbf{97.44}$ & $\mathbf{94.27}$\tabularnewline
\textit{ DGL2 $\lambda =$ 0.0001 $\phi =$ 0.01} & $97.93$ & $96.95$ & $95.60$ & $53.29$ & $11.90$\tabularnewline
\textit{ DGL2 $\lambda =$ 0.0001 $\phi =$ 0.05} & $98.29$ & $97.38$ & $96.06$ & $60.37$ & $13.68$\tabularnewline
\textit{ DGL2 $\lambda =$ 0.0001 $\phi =$ 0.1} & $98.02$ & $96.97$ & $95.66$ & $53.58$ & $12.26$\tabularnewline
\textit{ DGL2 $\lambda =$ 0.001 $\phi =$ 0.01} & $97.97$ & $97.00$ & $95.64$ & $54.55$ & $12.07$\tabularnewline
\textit{ DGL2 $\lambda =$ 0.001 $\phi =$ 0.05} & $98.20$ & $97.08$ & $95.63$ & $55.43$ & $12.53$\tabularnewline
\textit{ DGL2 $\lambda =$ 0.001 $\phi =$ 0.1} & $98.26$ & $97.42$ & $96.17$ & $60.86$ & $13.18$\tabularnewline
\textit{ DGL2 $\lambda =$ 0.01 $\phi =$ 0.01} & $98.20$ & $97.49$ & $96.60$ & $72.71$ & $19.84$\tabularnewline
\textit{ DGL2 $\lambda =$ 0.01 $\phi =$ 0.05} & $98.31$ & $97.60$ & $96.50$ & $69.66$ & $17.97$\tabularnewline
\textit{ DGL2 $\lambda =$ 0.01 $\phi =$ 0.1} & $98.30$ & $97.54$ & $96.40$ & $68.05$ & $17.61$\tabularnewline
\textit{ L1 $\lambda =$ 0.0001} & $98.15$ & $97.48$ & $96.14$ & $59.65$ & $13.19$\tabularnewline
\textit{ L1 $\lambda =$ 0.001} & $98.41$ & $97.69$ & $96.77$ & $74.96$ & $23.28$\tabularnewline
\textit{ L1 $\lambda =$ 0.01} & $97.73$ & $97.38$ & $97.12$ & $91.08$ & $74.35$\tabularnewline
\textit{ L1 $\lambda =$ 0.1} & $93.90$ & $93.38$ & $92.91$ & $86.62$ & $72.01$\tabularnewline
\textit{ L2 $\lambda =$ 0.0001} & $98.00$ & $97.05$ & $95.86$ & $57.90$ & $13.93$\tabularnewline
\textit{ L2 $\lambda =$ 0.001} & $97.88$ & $96.84$ & $95.39$ & $53.58$ & $12.41$\tabularnewline
\textit{ L2 $\lambda =$ 0.01} & $98.45$ & $98.00$ & $97.48$ & $84.87$ & $42.00$\tabularnewline
\textit{ L2 $\lambda =$ 0.1} & $98.12$ & $97.85$ & $97.53$ & $91.29$ & $69.87$\tabularnewline
\textit{ Rect} & $97.99$ & $96.80$ & $95.01$ & $49.06$ & $10.83$\tabularnewline
\textit{MT-DGL2 $\lambda =$ 0.001 $\phi =$ 0.01 } & $98.38$ & $98.00$ & $97.49$ & $81.37$ & $33.88$\tabularnewline
\textit{MT-DGL2 $\lambda =$ 0.001 $\phi =$ 0.05 } & $98.50$ & $98.06$ & $97.44$ & $81.52$ & $33.55$\tabularnewline
\textit{MT-DGL2 $\lambda =$ 0.001 $\phi =$ 0.1 } & $98.35$ & $97.87$ & $97.24$ & $79.36$ & $33.47$\tabularnewline
\textit{MT-DGL2 $\lambda =$ 0.01 $\phi =$ 0.01 } & $98.57$ & $98.26$ & $97.83$ & $89.19$ & $52.96$\tabularnewline
\textit{MT-DGL2 $\lambda =$ 0.01 $\phi =$ 0.05 } & $98.65$ & $98.26$ & $97.80$ & $89.69$ & $55.09$\tabularnewline
\textit{MT-DGL2 $\lambda =$ 0.01 $\phi =$ 0.1 } & $98.59$ & $98.31$ & $97.98$ & $90.03$ & $56.43$\tabularnewline
\textit{MT-DGL2 $\lambda =$ 0.1 $\phi =$ 0.01 } & $98.85$ & $98.74$ & $98.53$ & $95.75$ & $84.42$\tabularnewline
\textit{MT-DGL2 $\lambda =$ 0.1 $\phi =$ 0.05 } & $98.76$ & $98.59$ & $98.39$ & $95.16$ & $82.86$\tabularnewline
\textit{MT-DGL2 $\lambda =$ 0.1 $\phi =$ 0.1 } & $98.70$ & $98.54$ & $98.39$ & $95.37$ & $83.70$\tabularnewline
\textit{MT-DGL1 $\lambda =$ 0.001 $\phi =$ 0.01 } & $98.58$ & $98.38$ & $98.04$ & $90.46$ & $58.06$\tabularnewline
\textit{MT-DGL1 $\lambda =$ 0.001 $\phi =$ 0.05 } & $98.77$ & $98.66$ & $98.51$ & $96.08$ & $87.14$\tabularnewline
\textit{MT-DGL1 $\lambda =$ 0.001 $\phi =$ 0.1 } & $98.63$ & $98.47$ & $98.24$ & $93.36$ & $76.71$\tabularnewline
\textit{MT-DGL1 $\lambda =$ 0.01 $\phi =$ 0.01 } & $98.82$ & $98.69$ & $98.40$ & $95.79$ & $85.41$\tabularnewline
\textit{MT-DGL1 $\lambda =$ 0.01 $\phi =$ 0.05 } & $98.91$ & $98.82$ & $98.70$ & $97.16$ & $92.72$\tabularnewline
\textit{MT-DGL1 $\lambda =$ 0.01 $\phi =$ 0.1 } & $98.99$ & $98.94$ & $98.84$ & $97.44$ & $93.79$\tabularnewline
\textit{MT-DGL1 $\lambda =$ 0.1 $\phi =$ 0.01 } & $98.63$ & $98.50$ & $98.38$ & $97.04$ & $93.79$\tabularnewline
\textit{MT-DGL1 $\lambda =$ 0.1 $\phi =$ 0.05 } & $98.99$ & $98.97$ & $98.90$ & $98.25$ & $96.68$\tabularnewline
\textit{MT-DGL1 $\lambda =$ 0.1 $\phi =$ 0.1 } & $\mathbf{99.03}$ & $\mathbf{98.98}$ & $\mathbf{98.95}$ & $\mathbf{98.40}$ & $\mathbf{97.50}$\tabularnewline
\textit{MT } & $98.00$ & $97.20$ & $95.99$ & $63.38$ & $17.20$\tabularnewline

\hline
\end{tabular}
}
\end{table}

\begin{table}[!h]
\caption{Comparative results where all models are attacked by the L1-regularized rectifier network, \emph{L1}, using Laplacian ($\reg_1$) adversarial noise..}
\label{l1_attack}
\centering
\resizebox{12.5cm}{!} {%
\begin{tabular}{lrrrrr}
\multicolumn{1}{l}{\begin{tabular}[x]{@{}c@{}}\textbf{Model}\\\end{tabular}}&\multicolumn{1}{c}{\begin{tabular}[x]{@{}c@{}}\textbf{$\phi$ = 0.0}\\\end{tabular}}&\multicolumn{1}{c}{\begin{tabular}[x]{@{}c@{}}\textbf{$\phi$ = 0.005}\\\end{tabular}}&\multicolumn{1}{c}{\begin{tabular}[x]{@{}c@{}}\textbf{$\phi$ = 0.01}\\\end{tabular}}&\multicolumn{1}{c}{\begin{tabular}[x]{@{}c@{}}\textbf{$\phi$ = 0.05}\\\end{tabular}}&\multicolumn{1}{c}{\begin{tabular}[x]{@{}c@{}}\textbf{$\phi$ = 0.1}\\\end{tabular}}\tabularnewline
\hline
\textit{ DGL1 $\lambda =$ 0.0001 $\phi =$ 0.01} & $98.25$ & $97.31$ & $95.96$ & $56.68$ & $23.68$\tabularnewline
\textit{ DGL1 $\lambda =$ 0.0001 $\phi =$ 0.05} & $98.57$ & $98.15$ & $97.68$ & $88.75$ & $58.32$\tabularnewline
\textit{ DGL1 $\lambda =$ 0.0001 $\phi =$ 0.1} & $98.47$ & $98.11$ & $97.75$ & $89.65$ & $68.44$\tabularnewline
\textit{ DGL1 $\lambda =$ 0.001 $\phi =$ 0.01} & $98.03$ & $97.27$ & $96.12$ & $65.06$ & $26.64$\tabularnewline
\textit{ DGL1 $\lambda =$ 0.001 $\phi =$ 0.05} & $98.70$ & $98.41$ & $98.06$ & $92.78$ & $74.48$\tabularnewline
\textit{ DGL1 $\lambda =$ 0.001 $\phi =$ 0.1} & $98.62$ & $98.37$ & $98.12$ & $93.52$ & $82.12$\tabularnewline
\textit{ DGL1 $\lambda =$ 0.01 $\phi =$ 0.01} & $98.36$ & $97.93$ & $97.34$ & $87.04$ & $54.26$\tabularnewline
\textit{ DGL1 $\lambda =$ 0.01 $\phi =$ 0.05} & $98.62$ & $98.51$ & $98.36$ & $95.00$ & $85.33$\tabularnewline
\textit{ DGL1 $\lambda =$ 0.01 $\phi =$ 0.1} & $\mathbf{98.83}$ & $\mathbf{98.77}$ & $\mathbf{98.68}$ & $\mathbf{97.47}$ & $\mathbf{94.20}$\tabularnewline
\textit{ DGL2 $\lambda =$ 0.0001 $\phi =$ 0.01} & $97.93$ & $97.00$ & $95.72$ & $56.55$ & $23.07$\tabularnewline
\textit{ DGL2 $\lambda =$ 0.0001 $\phi =$ 0.05} & $98.29$ & $97.32$ & $95.94$ & $57.76$ & $23.86$\tabularnewline
\textit{ DGL2 $\lambda =$ 0.0001 $\phi =$ 0.1} & $98.02$ & $97.03$ & $95.74$ & $57.57$ & $23.96$\tabularnewline
\textit{ DGL2 $\lambda =$ 0.001 $\phi =$ 0.01} & $97.97$ & $96.98$ & $95.70$ & $57.95$ & $23.37$\tabularnewline
\textit{ DGL2 $\lambda =$ 0.001 $\phi =$ 0.05} & $98.20$ & $97.14$ & $95.71$ & $58.54$ & $24.01$\tabularnewline
\textit{ DGL2 $\lambda =$ 0.001 $\phi =$ 0.1} & $98.26$ & $97.36$ & $96.08$ & $57.30$ & $23.43$\tabularnewline
\textit{ DGL2 $\lambda =$ 0.01 $\phi =$ 0.01} & $98.20$ & $97.43$ & $96.48$ & $69.58$ & $28.98$\tabularnewline
\textit{ DGL2 $\lambda =$ 0.01 $\phi =$ 0.05} & $98.31$ & $97.57$ & $96.41$ & $66.05$ & $26.99$\tabularnewline
\textit{ DGL2 $\lambda =$ 0.01 $\phi =$ 0.1} & $98.30$ & $97.51$ & $96.32$ & $65.38$ & $26.63$\tabularnewline
\textit{ L1 $\lambda =$ 0.0001} & $98.15$ & $97.24$ & $95.61$ & $51.91$ & $21.44$\tabularnewline
\textit{ L1 $\lambda =$ 0.001} & $98.41$ & $97.62$ & $96.68$ & $71.69$ & $30.05$\tabularnewline
\textit{ L1 $\lambda =$ 0.01} & $97.73$ & $97.39$ & $97.08$ & $91.00$ & $73.68$\tabularnewline
\textit{ L1 $\lambda =$ 0.1} & $93.90$ & $93.38$ & $92.90$ & $86.96$ & $73.77$\tabularnewline
\textit{ L2 $\lambda =$ 0.0001} & $98.00$ & $97.03$ & $95.72$ & $56.20$ & $24.11$\tabularnewline
\textit{ L2 $\lambda =$ 0.001} & $97.88$ & $96.89$ & $95.50$ & $57.57$ & $23.54$\tabularnewline
\textit{ L2 $\lambda =$ 0.01} & $98.45$ & $97.98$ & $97.43$ & $83.01$ & $42.66$\tabularnewline
\textit{ L2 $\lambda =$ 0.1} & $98.12$ & $97.83$ & $97.52$ & $90.66$ & $68.11$\tabularnewline
\textit{ Rect} & $97.99$ & $96.98$ & $95.69$ & $58.53$ & $23.58$\tabularnewline
\textit{MT-DGL2 $\lambda =$ 0.001 $\phi =$ 0.01 } & $98.38$ & $97.99$ & $97.47$ & $80.50$ & $39.72$\tabularnewline
\textit{MT-DGL2 $\lambda =$ 0.001 $\phi =$ 0.05 } & $98.50$ & $98.06$ & $97.44$ & $80.28$ & $39.47$\tabularnewline
\textit{MT-DGL2 $\lambda =$ 0.001 $\phi =$ 0.1 } & $98.35$ & $97.89$ & $97.27$ & $78.97$ & $39.08$\tabularnewline
\textit{MT-DGL2 $\lambda =$ 0.01 $\phi =$ 0.01 } & $98.57$ & $98.28$ & $97.84$ & $88.89$ & $55.29$\tabularnewline
\textit{MT-DGL2 $\lambda =$ 0.01 $\phi =$ 0.05 } & $98.65$ & $98.23$ & $97.81$ & $89.46$ & $56.76$\tabularnewline
\textit{MT-DGL2 $\lambda =$ 0.01 $\phi =$ 0.1 } & $98.59$ & $98.35$ & $97.98$ & $89.58$ & $58.33$\tabularnewline
\textit{MT-DGL2 $\lambda =$ 0.1 $\phi =$ 0.01 } & $98.85$ & $98.74$ & $98.57$ & $95.67$ & $84.08$\tabularnewline
\textit{MT-DGL2 $\lambda =$ 0.1 $\phi =$ 0.05 } & $98.76$ & $98.57$ & $98.36$ & $95.12$ & $81.97$\tabularnewline
\textit{MT-DGL2 $\lambda =$ 0.1 $\phi =$ 0.1 } & $98.70$ & $98.55$ & $98.38$ & $95.30$ & $82.72$\tabularnewline
\textit{MT-DGL1 $\lambda =$ 0.001 $\phi =$ 0.01 } & $98.58$ & $98.36$ & $98.06$ & $90.06$ & $59.26$\tabularnewline
\textit{MT-DGL1 $\lambda =$ 0.001 $\phi =$ 0.05 } & $98.77$ & $98.64$ & $98.48$ & $95.92$ & $86.91$\tabularnewline
\textit{MT-DGL1 $\lambda =$ 0.001 $\phi =$ 0.1 } & $98.63$ & $98.47$ & $98.27$ & $93.21$ & $76.36$\tabularnewline
\textit{MT-DGL1 $\lambda =$ 0.01 $\phi =$ 0.01 } & $98.82$ & $98.66$ & $98.40$ & $95.73$ & $85.10$\tabularnewline
\textit{MT-DGL1 $\lambda =$ 0.01 $\phi =$ 0.05 } & $98.91$ & $98.79$ & $98.70$ & $97.18$ & $92.77$\tabularnewline
\textit{MT-DGL1 $\lambda =$ 0.01 $\phi =$ 0.1 } & $98.99$ & $98.94$ & $98.78$ & $97.45$ & $93.90$\tabularnewline
\textit{MT-DGL1 $\lambda =$ 0.1 $\phi =$ 0.01 } & $98.63$ & $98.51$ & $98.40$ & $96.99$ & $93.65$\tabularnewline
\textit{MT-DGL1 $\lambda =$ 0.1 $\phi =$ 0.05 } & $98.99$ & $98.96$ & $98.89$ & $98.26$ & $96.68$\tabularnewline
\textit{MT-DGL1 $\lambda =$ 0.1 $\phi =$ 0.1 } & $\mathbf{99.03}$ & $\mathbf{98.97}$ & $\mathbf{98.93}$ & $\mathbf{98.37}$ & $\mathbf{97.33}$\tabularnewline
\textit{MT } & $98.00$ & $97.18$ & $95.86$ & $63.03$ & $25.51$\tabularnewline
\hline
\end{tabular}
}
\end{table}

\begin{table}[!h]
\caption{Comparative results where all models are attacked by the L2-regularized rectifier network, \emph{L2}, using Laplacian ($\reg_1$) adversarial noise.}
\label{l2_attack}
\centering
\resizebox{12.5cm}{!} {%
\begin{tabular}{lrrrrr}
\multicolumn{1}{l}{\begin{tabular}[x]{@{}c@{}}\textbf{Model}\\\end{tabular}}&\multicolumn{1}{c}{\begin{tabular}[x]{@{}c@{}}\textbf{$\phi$ = 0.0}\\\end{tabular}}&\multicolumn{1}{c}{\begin{tabular}[x]{@{}c@{}}\textbf{$\phi$ = 0.005}\\\end{tabular}}&\multicolumn{1}{c}{\begin{tabular}[x]{@{}c@{}}\textbf{$\phi$ = 0.01}\\\end{tabular}}&\multicolumn{1}{c}{\begin{tabular}[x]{@{}c@{}}\textbf{$\phi$ = 0.05}\\\end{tabular}}&\multicolumn{1}{c}{\begin{tabular}[x]{@{}c@{}}\textbf{$\phi$ = 0.1}\\\end{tabular}}\tabularnewline
\hline
\textit{ DGL1 $\lambda =$ 0.0001 $\phi =$ 0.01} & $98.25$ & $97.70$ & $96.81$ & $75.66$ & $25.26$\tabularnewline
\textit{ DGL1 $\lambda =$ 0.0001 $\phi =$ 0.05} & $98.57$ & $98.28$ & $97.99$ & $93.04$ & $73.97$\tabularnewline
\textit{ DGL1 $\lambda =$ 0.0001 $\phi =$ 0.1} & $98.47$ & $98.26$ & $97.92$ & $93.39$ & $77.97$\tabularnewline
\textit{ DGL1 $\lambda =$ 0.001 $\phi =$ 0.01} & $98.03$ & $97.48$ & $96.94$ & $81.26$ & $32.74$\tabularnewline
\textit{ DGL1 $\lambda =$ 0.001 $\phi =$ 0.05} & $98.70$ & $98.48$ & $98.22$ & $94.82$ & $82.96$\tabularnewline
\textit{ DGL1 $\lambda =$ 0.001 $\phi =$ 0.1} & $98.62$ & $98.44$ & $98.27$ & $95.04$ & $86.12$\tabularnewline
\textit{ DGL1 $\lambda =$ 0.01 $\phi =$ 0.01} & $98.36$ & $98.04$ & $97.52$ & $90.49$ & $63.92$\tabularnewline
\textit{ DGL1 $\lambda =$ 0.01 $\phi =$ 0.05} & $98.62$ & $98.51$ & $98.42$ & $95.57$ & $87.67$\tabularnewline
\textit{ DGL1 $\lambda =$ 0.01 $\phi =$ 0.1} & $\mathbf{98.83}$ & $\mathbf{98.77}$ & $\mathbf{98.67}$ & $\mathbf{97.44}$ & $\mathbf{94.27}$\tabularnewline
\textit{ DGL2 $\lambda =$ 0.0001 $\phi =$ 0.01} & $97.93$ & $97.28$ & $96.50$ & $76.93$ & $25.90$\tabularnewline
\textit{ DGL2 $\lambda =$ 0.0001 $\phi =$ 0.05} & $98.29$ & $97.68$ & $96.75$ & $75.97$ & $24.43$\tabularnewline
\textit{ DGL2 $\lambda =$ 0.0001 $\phi =$ 0.1} & $98.02$ & $97.35$ & $96.57$ & $76.87$ & $26.25$\tabularnewline
\textit{ DGL2 $\lambda =$ 0.001 $\phi =$ 0.01} & $97.97$ & $97.38$ & $96.46$ & $77.19$ & $26.88$\tabularnewline
\textit{ DGL2 $\lambda =$ 0.001 $\phi =$ 0.05} & $98.20$ & $97.52$ & $96.66$ & $77.90$ & $26.53$\tabularnewline
\textit{ DGL2 $\lambda =$ 0.001 $\phi =$ 0.1} & $98.26$ & $97.72$ & $96.89$ & $77.18$ & $25.03$\tabularnewline
\textit{ DGL2 $\lambda =$ 0.01 $\phi =$ 0.01} & $98.20$ & $97.76$ & $97.02$ & $83.13$ & $36.57$\tabularnewline
\textit{ DGL2 $\lambda =$ 0.01 $\phi =$ 0.05} & $98.31$ & $97.78$ & $97.14$ & $82.35$ & $34.70$\tabularnewline
\textit{ DGL2 $\lambda =$ 0.01 $\phi =$ 0.1} & $98.30$ & $97.77$ & $97.17$ & $82.02$ & $33.80$\tabularnewline
\textit{ L1 $\lambda =$ 0.0001} & $98.15$ & $97.63$ & $96.91$ & $75.97$ & $24.07$\tabularnewline
\textit{ L1 $\lambda =$ 0.001} & $98.41$ & $97.71$ & $96.97$ & $76.47$ & $24.02$\tabularnewline
\textit{ L1 $\lambda =$ 0.01} & $97.73$ & $97.32$ & $96.97$ & $89.07$ & $64.45$\tabularnewline
\textit{ L1 $\lambda =$ 0.1} & $93.90$ & $93.40$ & $92.83$ & $86.17$ & $68.98$\tabularnewline
\textit{ L2 $\lambda =$ 0.0001} & $98.00$ & $97.37$ & $96.59$ & $76.52$ & $26.14$\tabularnewline
\textit{ L2 $\lambda =$ 0.001} & $97.88$ & $97.22$ & $96.44$ & $76.19$ & $25.73$\tabularnewline
\textit{ L2 $\lambda =$ 0.01} & $98.45$ & $97.74$ & $96.52$ & $60.93$ & $14.34$\tabularnewline
\textit{ L2 $\lambda =$ 0.1} & $98.12$ & $97.81$ & $97.19$ & $86.46$ & $47.90$\tabularnewline
\textit{ Rect} & $97.99$ & $97.32$ & $96.57$ & $77.87$ & $26.99$\tabularnewline
\textit{MT-DGL2 $\lambda =$ 0.001 $\phi =$ 0.01 } & $98.38$ & $98.07$ & $97.63$ & $86.65$ & $45.44$\tabularnewline
\textit{MT-DGL2 $\lambda =$ 0.001 $\phi =$ 0.05 } & $98.50$ & $98.15$ & $97.69$ & $86.91$ & $44.76$\tabularnewline
\textit{MT-DGL2 $\lambda =$ 0.001 $\phi =$ 0.1 } & $98.35$ & $97.99$ & $97.58$ & $86.04$ & $43.86$\tabularnewline
\textit{MT-DGL2 $\lambda =$ 0.01 $\phi =$ 0.01 } & $98.57$ & $98.33$ & $97.96$ & $91.69$ & $63.27$\tabularnewline
\textit{MT-DGL2 $\lambda =$ 0.01 $\phi =$ 0.05 } & $98.65$ & $98.33$ & $98.01$ & $92.24$ & $65.58$\tabularnewline
\textit{MT-DGL2 $\lambda =$ 0.01 $\phi =$ 0.1 } & $98.59$ & $98.41$ & $98.10$ & $92.37$ & $66.75$\tabularnewline
\textit{MT-DGL2 $\lambda =$ 0.1 $\phi =$ 0.01 } & $98.85$ & $98.76$ & $98.59$ & $96.03$ & $85.26$\tabularnewline
\textit{MT-DGL2 $\lambda =$ 0.1 $\phi =$ 0.05 } & $98.76$ & $98.64$ & $98.43$ & $95.71$ & $84.42$\tabularnewline
\textit{MT-DGL2 $\lambda =$ 0.1 $\phi =$ 0.1 } & $98.70$ & $98.56$ & $98.42$ & $95.78$ & $85.20$\tabularnewline
\textit{MT-DGL1 $\lambda =$ 0.001 $\phi =$ 0.01 } & $98.58$ & $98.41$ & $98.16$ & $92.21$ & $65.98$\tabularnewline
\textit{MT-DGL1 $\lambda =$ 0.001 $\phi =$ 0.05 } & $98.77$ & $98.65$ & $98.53$ & $96.28$ & $88.39$\tabularnewline
\textit{MT-DGL1 $\lambda =$ 0.001 $\phi =$ 0.1 } & $98.63$ & $98.48$ & $98.33$ & $94.49$ & $79.56$\tabularnewline
\textit{MT-DGL1 $\lambda =$ 0.01 $\phi =$ 0.01 } & $98.82$ & $98.69$ & $98.45$ & $96.13$ & $86.22$\tabularnewline
\textit{MT-DGL1 $\lambda =$ 0.01 $\phi =$ 0.05 } & $98.91$ & $98.82$ & $98.71$ & $97.26$ & $92.70$\tabularnewline
\textit{MT-DGL1 $\lambda =$ 0.01 $\phi =$ 0.1 } & $98.99$ & $98.94$ & $98.81$ & $97.43$ & $94.04$\tabularnewline
\textit{MT-DGL1 $\lambda =$ 0.1 $\phi =$ 0.01 } & $98.63$ & $98.51$ & $98.41$ & $96.98$ & $93.27$\tabularnewline
\textit{MT-DGL1 $\lambda =$ 0.1 $\phi =$ 0.05 } & $98.99$ & $98.96$ & $98.90$ & $98.24$ & $96.58$\tabularnewline
\textit{MT-DGL1 $\lambda =$ 0.1 $\phi =$ 0.1 } & $\mathbf{99.03}$ & $\mathbf{98.98}$ & $\mathbf{98.91}$ & $\mathbf{98.36}$ & $\mathbf{97.37}$\tabularnewline
\textit{MT } & $98.00$ & $97.37$ & $96.59$ & $74.09$ & $23.44$\tabularnewline

\hline
\end{tabular}
}
\end{table}

\begin{table}[!h]
\caption{Comparative results where all models are attacked by the DataGrad-L1-regularized rectifier network, \emph{DGL1}, using Laplacian ($\reg_1$) adversarial noise.}
\label{dgl1_attack}
\centering
\resizebox{12.5cm}{!} {%
\begin{tabular}{lrrrrr}
\multicolumn{1}{l}{\begin{tabular}[x]{@{}c@{}}\textbf{Model}\\\end{tabular}}&\multicolumn{1}{c}{\begin{tabular}[x]{@{}c@{}}\textbf{$\phi$ = 0.0}\\\end{tabular}}&\multicolumn{1}{c}{\begin{tabular}[x]{@{}c@{}}\textbf{$\phi$ = 0.005}\\\end{tabular}}&\multicolumn{1}{c}{\begin{tabular}[x]{@{}c@{}}\textbf{$\phi$ = 0.01}\\\end{tabular}}&\multicolumn{1}{c}{\begin{tabular}[x]{@{}c@{}}\textbf{$\phi$ = 0.05}\\\end{tabular}}&\multicolumn{1}{c}{\begin{tabular}[x]{@{}c@{}}\textbf{$\phi$ = 0.1}\\\end{tabular}}\tabularnewline
\hline
\textit{ DGL1 $\lambda =$ 0.0001 $\phi =$ 0.01} & $98.25$ & $98.05$ & $97.71$ & $93.34$ & $78.95$\tabularnewline
\textit{ DGL1 $\lambda =$ 0.0001 $\phi =$ 0.05} & $98.57$ & $98.31$ & $98.14$ & $94.85$ & $83.99$\tabularnewline
\textit{ DGL1 $\lambda =$ 0.0001 $\phi =$ 0.1} & $98.47$ & $98.27$ & $98.05$ & $94.89$ & $84.31$\tabularnewline
\textit{ DGL1 $\lambda =$ 0.001 $\phi =$ 0.01} & $98.03$ & $97.75$ & $97.42$ & $92.87$ & $78.07$\tabularnewline
\textit{ DGL1 $\lambda =$ 0.001 $\phi =$ 0.05} & $98.70$ & $98.47$ & $98.21$ & $94.83$ & $84.16$\tabularnewline
\textit{ DGL1 $\lambda =$ 0.001 $\phi =$ 0.1} & $\mathbf{98.62}$ & $\mathbf{98.42}$ & $\mathbf{98.21}$ & $\mathbf{94.73}$ & $\mathbf{84.87}$\tabularnewline
\textit{ DGL1 $\lambda =$ 0.01 $\phi =$ 0.01} & $98.36$ & $98.15$ & $97.83$ & $93.55$ & $80.92$\tabularnewline
\textit{ DGL1 $\lambda =$ 0.01 $\phi =$ 0.05} & $98.62$ & $98.49$ & $98.33$ & $94.34$ & $82.77$\tabularnewline
\textit{ DGL1 $\lambda =$ 0.01 $\phi =$ 0.1} & $98.83$ & $98.63$ & $98.34$ & $94.41$ & $83.74$\tabularnewline
\textit{ DGL2 $\lambda =$ 0.0001 $\phi =$ 0.01} & $97.93$ & $97.71$ & $97.31$ & $92.55$ & $77.17$\tabularnewline
\textit{ DGL2 $\lambda =$ 0.0001 $\phi =$ 0.05} & $98.29$ & $97.99$ & $97.72$ & $93.20$ & $78.77$\tabularnewline
\textit{ DGL2 $\lambda =$ 0.0001 $\phi =$ 0.1} & $98.02$ & $97.73$ & $97.39$ & $92.77$ & $77.69$\tabularnewline
\textit{ DGL2 $\lambda =$ 0.001 $\phi =$ 0.01} & $97.97$ & $97.64$ & $97.37$ & $92.60$ & $76.74$\tabularnewline
\textit{ DGL2 $\lambda =$ 0.001 $\phi =$ 0.05} & $98.20$ & $97.95$ & $97.52$ & $92.97$ & $77.97$\tabularnewline
\textit{ DGL2 $\lambda =$ 0.001 $\phi =$ 0.1} & $98.26$ & $98.00$ & $97.73$ & $93.44$ & $78.81$\tabularnewline
\textit{ DGL2 $\lambda =$ 0.01 $\phi =$ 0.01} & $98.20$ & $97.95$ & $97.65$ & $93.26$ & $79.59$\tabularnewline
\textit{ DGL2 $\lambda =$ 0.01 $\phi =$ 0.05} & $98.31$ & $98.07$ & $97.69$ & $93.14$ & $78.58$\tabularnewline
\textit{ DGL2 $\lambda =$ 0.01 $\phi =$ 0.1} & $98.30$ & $98.02$ & $97.69$ & $92.98$ & $77.97$\tabularnewline
\textit{ L1 $\lambda =$ 0.0001} & $98.15$ & $97.89$ & $97.61$ & $93.22$ & $78.80$\tabularnewline
\textit{ L1 $\lambda =$ 0.001} & $98.41$ & $98.10$ & $97.83$ & $94.07$ & $81.43$\tabularnewline
\textit{ L1 $\lambda =$ 0.01} & $97.73$ & $97.46$ & $97.22$ & $93.56$ & $83.95$\tabularnewline
\textit{ L1 $\lambda =$ 0.1} & $93.90$ & $93.56$ & $93.29$ & $89.81$ & $81.29$\tabularnewline
\textit{ L2 $\lambda =$ 0.0001} & $98.00$ & $97.76$ & $97.41$ & $92.57$ & $77.57$\tabularnewline
\textit{ L2 $\lambda =$ 0.001} & $97.88$ & $97.59$ & $97.28$ & $92.37$ & $76.68$\tabularnewline
\textit{ L2 $\lambda =$ 0.01} & $98.45$ & $98.24$ & $98.03$ & $94.26$ & $82.19$\tabularnewline
\textit{ L2 $\lambda =$ 0.1} & $98.12$ & $97.96$ & $97.75$ & $94.44$ & $84.35$\tabularnewline
\textit{ Rect} & $97.99$ & $97.68$ & $97.34$ & $92.89$ & $78.44$\tabularnewline
\textit{MT-DGL2 $\lambda =$ 0.001 $\phi =$ 0.01 } & $98.38$ & $98.15$ & $97.94$ & $93.69$ & $79.16$\tabularnewline
\textit{MT-DGL2 $\lambda =$ 0.001 $\phi =$ 0.05 } & $98.50$ & $98.24$ & $98.00$ & $93.77$ & $79.74$\tabularnewline
\textit{MT-DGL2 $\lambda =$ 0.001 $\phi =$ 0.1 } & $98.35$ & $98.10$ & $97.87$ & $93.41$ & $77.41$\tabularnewline
\textit{MT-DGL2 $\lambda =$ 0.01 $\phi =$ 0.01 } & $98.57$ & $98.38$ & $98.16$ & $94.37$ & $81.79$\tabularnewline
\textit{MT-DGL2 $\lambda =$ 0.01 $\phi =$ 0.05 } & $98.65$ & $98.40$ & $98.13$ & $94.53$ & $81.87$\tabularnewline
\textit{MT-DGL2 $\lambda =$ 0.01 $\phi =$ 0.1 } & $98.59$ & $98.45$ & $98.19$ & $94.50$ & $82.22$\tabularnewline
\textit{MT-DGL2 $\lambda =$ 0.1 $\phi =$ 0.01 } & $98.85$ & $98.75$ & $98.58$ & $96.10$ & $86.59$\tabularnewline
\textit{MT-DGL2 $\lambda =$ 0.1 $\phi =$ 0.05 } & $98.76$ & $98.58$ & $98.35$ & $95.58$ & $85.48$\tabularnewline
\textit{MT-DGL2 $\lambda =$ 0.1 $\phi =$ 0.1 } & $98.70$ & $98.54$ & $98.42$ & $95.72$ & $85.46$\tabularnewline
\textit{MT-DGL1 $\lambda =$ 0.001 $\phi =$ 0.01 } & $98.58$ & $98.50$ & $98.25$ & $94.80$ & $83.03$\tabularnewline
\textit{MT-DGL1 $\lambda =$ 0.001 $\phi =$ 0.05 } & $98.77$ & $98.64$ & $98.47$ & $95.99$ & $88.08$\tabularnewline
\textit{MT-DGL1 $\lambda =$ 0.001 $\phi =$ 0.1 } & $98.63$ & $98.48$ & $98.32$ & $95.33$ & $85.11$\tabularnewline
\textit{MT-DGL1 $\lambda =$ 0.01 $\phi =$ 0.01 } & $98.82$ & $98.66$ & $98.43$ & $95.68$ & $86.38$\tabularnewline
\textit{MT-DGL1 $\lambda =$ 0.01 $\phi =$ 0.05 } & $98.91$ & $98.79$ & $98.59$ & $96.61$ & $90.05$\tabularnewline
\textit{MT-DGL1 $\lambda =$ 0.01 $\phi =$ 0.1 } & $98.99$ & $98.91$ & $98.77$ & $96.67$ & $90.82$\tabularnewline
\textit{MT-DGL1 $\lambda =$ 0.1 $\phi =$ 0.01 } & $98.63$ & $98.50$ & $98.32$ & $96.43$ & $91.62$\tabularnewline
\textit{MT-DGL1 $\lambda =$ 0.1 $\phi =$ 0.05 } & $98.99$ & $98.92$ & $98.86$ & $97.90$ & $95.35$\tabularnewline
\textit{MT-DGL1 $\lambda =$ 0.1 $\phi =$ 0.1 } & $\mathbf{99.03}$ & $\mathbf{98.98}$ & $\mathbf{98.87}$ & $\mathbf{98.08}$ & $\mathbf{96.42}$\tabularnewline
\textit{MT } & $98.00$ & $97.65$ & $97.30$ & $92.13$ & $75.10$\tabularnewline
\hline
\end{tabular}
}
\end{table}

\begin{table}[!h]
\caption{Comparative results where all models are attacked by the DataGrad-L2-regularized rectifier network, \emph{DGL2}, using Laplacian ($\reg_1$) adversarial noise.}
\label{dgl2_attack}
\centering
\resizebox{12.5cm}{!} {%
\begin{tabular}{lrrrrr}
\multicolumn{1}{l}{\begin{tabular}[x]{@{}c@{}}\textbf{Model}\\\end{tabular}}&\multicolumn{1}{c}{\begin{tabular}[x]{@{}c@{}}\textbf{$\phi$ = 0.0}\\\end{tabular}}&\multicolumn{1}{c}{\begin{tabular}[x]{@{}c@{}}\textbf{$\phi$ = 0.005}\\\end{tabular}}&\multicolumn{1}{c}{\begin{tabular}[x]{@{}c@{}}\textbf{$\phi$ = 0.01}\\\end{tabular}}&\multicolumn{1}{c}{\begin{tabular}[x]{@{}c@{}}\textbf{$\phi$ = 0.05}\\\end{tabular}}&\multicolumn{1}{c}{\begin{tabular}[x]{@{}c@{}}\textbf{$\phi$ = 0.1}\\\end{tabular}}\tabularnewline
\hline
\textit{ DGL1 $\lambda =$ 0.0001 $\phi =$ 0.01} & $98.25$ & $97.41$ & $96.16$ & $60.96$ & $14.33$\tabularnewline
\textit{ DGL1 $\lambda =$ 0.0001 $\phi =$ 0.05} & $98.57$ & $98.13$ & $97.67$ & $88.27$ & $54.88$\tabularnewline
\textit{ DGL1 $\lambda =$ 0.0001 $\phi =$ 0.1} & $98.47$ & $98.12$ & $97.66$ & $89.06$ & $65.93$\tabularnewline
\textit{ DGL1 $\lambda =$ 0.001 $\phi =$ 0.01} & $98.03$ & $97.28$ & $96.12$ & $67.04$ & $16.81$\tabularnewline
\textit{ DGL1 $\lambda =$ 0.001 $\phi =$ 0.05} & $98.70$ & $98.40$ & $97.95$ & $92.00$ & $70.31$\tabularnewline
\textit{ DGL1 $\lambda =$ 0.001 $\phi =$ 0.1} & $98.62$ & $98.36$ & $98.07$ & $92.92$ & $79.93$\tabularnewline
\textit{ DGL1 $\lambda =$ 0.01 $\phi =$ 0.01} & $98.36$ & $97.91$ & $97.28$ & $86.01$ & $48.12$\tabularnewline
\textit{ DGL1 $\lambda =$ 0.01 $\phi =$ 0.05} & $98.62$ & $98.50$ & $98.35$ & $94.81$ & $83.50$\tabularnewline
\textit{ DGL1 $\lambda =$ 0.01 $\phi =$ 0.1} & $\mathbf{98.83}$ & $\mathbf{98.77}$ & $\mathbf{98.67}$ & $\mathbf{97.28}$ & $\mathbf{93.75}$\tabularnewline
\textit{ DGL2 $\lambda =$ 0.0001 $\phi =$ 0.01} & $97.93$ & $97.05$ & $95.73$ & $59.03$ & $13.69$\tabularnewline
\textit{ DGL2 $\lambda =$ 0.0001 $\phi =$ 0.05} & $98.29$ & $97.38$ & $96.06$ & $60.67$ & $14.03$\tabularnewline
\textit{ DGL2 $\lambda =$ 0.0001 $\phi =$ 0.1} & $98.02$ & $97.11$ & $95.77$ & $60.25$ & $14.48$\tabularnewline
\textit{ DGL2 $\lambda =$ 0.001 $\phi =$ 0.01} & $97.97$ & $97.04$ & $95.85$ & $60.31$ & $14.08$\tabularnewline
\textit{ DGL2 $\lambda =$ 0.001 $\phi =$ 0.05} & $98.20$ & $97.17$ & $95.90$ & $61.10$ & $14.52$\tabularnewline
\textit{ DGL2 $\lambda =$ 0.001 $\phi =$ 0.1} & $98.26$ & $97.43$ & $96.16$ & $60.24$ & $13.34$\tabularnewline
\textit{ DGL2 $\lambda =$ 0.01 $\phi =$ 0.01} & $98.20$ & $97.41$ & $96.43$ & $69.33$ & $17.98$\tabularnewline
\textit{ DGL2 $\lambda =$ 0.01 $\phi =$ 0.05} & $98.31$ & $97.55$ & $96.26$ & $63.85$ & $14.71$\tabularnewline
\textit{ DGL2 $\lambda =$ 0.01 $\phi =$ 0.1} & $98.30$ & $97.34$ & $95.85$ & $57.83$ & $13.00$\tabularnewline
\textit{ L1 $\lambda =$ 0.0001} & $98.15$ & $97.48$ & $96.18$ & $61.12$ & $13.89$\tabularnewline
\textit{ L1 $\lambda =$ 0.001} & $98.41$ & $97.66$ & $96.82$ & $75.51$ & $23.36$\tabularnewline
\textit{ L1 $\lambda =$ 0.01} & $97.73$ & $97.34$ & $97.10$ & $91.06$ & $73.55$\tabularnewline
\textit{ L1 $\lambda =$ 0.1} & $93.90$ & $93.38$ & $92.94$ & $86.84$ & $72.99$\tabularnewline
\textit{ L2 $\lambda =$ 0.0001} & $98.00$ & $97.05$ & $95.82$ & $59.47$ & $14.10$\tabularnewline
\textit{ L2 $\lambda =$ 0.001} & $97.88$ & $96.93$ & $95.69$ & $59.73$ & $14.67$\tabularnewline
\textit{ L2 $\lambda =$ 0.01} & $98.45$ & $98.02$ & $97.51$ & $84.91$ & $42.24$\tabularnewline
\textit{ L2 $\lambda =$ 0.1} & $98.12$ & $97.84$ & $97.53$ & $91.02$ & $68.14$\tabularnewline
\textit{ Rect} & $97.99$ & $97.04$ & $95.78$ & $61.98$ & $15.02$\tabularnewline
\textit{MT-DGL2 $\lambda =$ 0.001 $\phi =$ 0.01 } & $98.38$ & $97.98$ & $97.42$ & $80.81$ & $32.25$\tabularnewline
\textit{MT-DGL2 $\lambda =$ 0.001 $\phi =$ 0.05 } & $98.50$ & $98.07$ & $97.41$ & $80.76$ & $32.50$\tabularnewline
\textit{MT-DGL2 $\lambda =$ 0.001 $\phi =$ 0.1 } & $98.35$ & $97.90$ & $97.30$ & $79.30$ & $32.06$\tabularnewline
\textit{MT-DGL2 $\lambda =$ 0.01 $\phi =$ 0.01 } & $98.57$ & $98.26$ & $97.83$ & $88.36$ & $50.17$\tabularnewline
\textit{MT-DGL2 $\lambda =$ 0.01 $\phi =$ 0.05 } & $98.65$ & $98.24$ & $97.77$ & $88.79$ & $51.86$\tabularnewline
\textit{MT-DGL2 $\lambda =$ 0.01 $\phi =$ 0.1 } & $98.59$ & $98.31$ & $97.93$ & $89.13$ & $53.25$\tabularnewline
\textit{MT-DGL2 $\lambda =$ 0.1 $\phi =$ 0.01 } & $98.85$ & $98.76$ & $98.53$ & $95.51$ & $82.51$\tabularnewline
\textit{MT-DGL2 $\lambda =$ 0.1 $\phi =$ 0.05 } & $98.76$ & $98.58$ & $98.33$ & $94.74$ & $80.04$\tabularnewline
\textit{MT-DGL2 $\lambda =$ 0.1 $\phi =$ 0.1 } & $98.70$ & $98.52$ & $98.34$ & $95.08$ & $81.28$\tabularnewline
\textit{MT-DGL1 $\lambda =$ 0.001 $\phi =$ 0.01 } & $98.58$ & $98.36$ & $98.06$ & $89.82$ & $55.87$\tabularnewline
\textit{MT-DGL1 $\lambda =$ 0.001 $\phi =$ 0.05 } & $98.77$ & $98.65$ & $98.49$ & $95.86$ & $85.98$\tabularnewline
\textit{MT-DGL1 $\lambda =$ 0.001 $\phi =$ 0.1 } & $98.63$ & $98.46$ & $98.24$ & $92.57$ & $74.01$\tabularnewline
\textit{MT-DGL1 $\lambda =$ 0.01 $\phi =$ 0.01 } & $98.82$ & $98.65$ & $98.41$ & $95.46$ & $83.91$\tabularnewline
\textit{MT-DGL1 $\lambda =$ 0.01 $\phi =$ 0.05 } & $98.91$ & $98.79$ & $98.66$ & $96.99$ & $92.26$\tabularnewline
\textit{MT-DGL1 $\lambda =$ 0.01 $\phi =$ 0.1 } & $98.99$ & $98.94$ & $98.80$ & $97.18$ & $93.14$\tabularnewline
\textit{MT-DGL1 $\lambda =$ 0.1 $\phi =$ 0.01 } & $98.63$ & $98.50$ & $98.37$ & $97.04$ & $93.33$\tabularnewline
\textit{MT-DGL1 $\lambda =$ 0.1 $\phi =$ 0.05 } & $98.99$ & $98.95$ & $98.89$ & $98.19$ & $96.31$\tabularnewline
\textit{MT-DGL1 $\lambda =$ 0.1 $\phi =$ 0.1 } & $\mathbf{99.03}$ & $\mathbf{98.98}$ & $\mathbf{98.94}$ & $\mathbf{98.31}$ & $\mathbf{97.36}$\tabularnewline
\textit{MT } & $98.00$ & $97.22$ & $96.04$ & $64.16$ & $17.38$\tabularnewline

\hline
\end{tabular}
}
\end{table}

\begin{table}[!h]
\caption{Comparative results where all models are attacked by the multi-task rectifier network, \emph{MT}, using Laplacian ($\reg_1$) adversarial noise driven by the target task (digit recognition).}
\label{mt_attack}
\centering
\resizebox{12.5cm}{!} {%
\begin{tabular}{lrrrrr}
\multicolumn{1}{l}{\begin{tabular}[x]{@{}c@{}}\textbf{Model}\\\end{tabular}}&\multicolumn{1}{c}{\begin{tabular}[x]{@{}c@{}}\textbf{$\phi$ = 0.0}\\\end{tabular}}&\multicolumn{1}{c}{\begin{tabular}[x]{@{}c@{}}\textbf{$\phi$ = 0.005}\\\end{tabular}}&\multicolumn{1}{c}{\begin{tabular}[x]{@{}c@{}}\textbf{$\phi$ = 0.01}\\\end{tabular}}&\multicolumn{1}{c}{\begin{tabular}[x]{@{}c@{}}\textbf{$\phi$ = 0.05}\\\end{tabular}}&\multicolumn{1}{c}{\begin{tabular}[x]{@{}c@{}}\textbf{$\phi$ = 0.1}\\\end{tabular}}\tabularnewline
\hline
\textit{ DGL1 $\lambda =$ 0.0001 $\phi =$ 0.01} & $98.25$ & $97.95$ & $97.50$ & $89.19$ & $57.88$\tabularnewline
\textit{ DGL1 $\lambda =$ 0.0001 $\phi =$ 0.05} & $98.57$ & $98.38$ & $98.16$ & $95.99$ & $89.33$\tabularnewline
\textit{ DGL1 $\lambda =$ 0.0001 $\phi =$ 0.1} & $98.47$ & $98.28$ & $98.19$ & $96.15$ & $89.95$\tabularnewline
\textit{ DGL1 $\lambda =$ 0.001 $\phi =$ 0.01} & $98.03$ & $97.71$ & $97.36$ & $90.86$ & $65.55$\tabularnewline
\textit{ DGL1 $\lambda =$ 0.001 $\phi =$ 0.05} & $98.70$ & $98.56$ & $98.41$ & $96.74$ & $92.11$\tabularnewline
\textit{ DGL1 $\lambda =$ 0.001 $\phi =$ 0.1} & $98.62$ & $98.52$ & $98.39$ & $96.94$ & $93.02$\tabularnewline
\textit{ DGL1 $\lambda =$ 0.01 $\phi =$ 0.01} & $98.36$ & $98.18$ & $97.99$ & $95.13$ & $86.87$\tabularnewline
\textit{ DGL1 $\lambda =$ 0.01 $\phi =$ 0.05} & $98.62$ & $98.54$ & $98.49$ & $97.09$ & $94.23$\tabularnewline
\textit{ DGL1 $\lambda =$ 0.01 $\phi =$ 0.1} & $\mathbf{98.83}$ & $\mathbf{98.80}$ & $\mathbf{98.75}$ & $\mathbf{98.12}$ & $\mathbf{96.73}$\tabularnewline
\textit{ DGL2 $\lambda =$ 0.0001 $\phi =$ 0.01} & $97.93$ & $97.50$ & $97.13$ & $88.86$ & $56.48$\tabularnewline
\textit{ DGL2 $\lambda =$ 0.0001 $\phi =$ 0.05} & $98.29$ & $97.91$ & $97.51$ & $89.36$ & $57.68$\tabularnewline
\textit{ DGL2 $\lambda =$ 0.0001 $\phi =$ 0.1} & $98.02$ & $97.59$ & $97.16$ & $88.85$ & $56.84$\tabularnewline
\textit{ DGL2 $\lambda =$ 0.001 $\phi =$ 0.01} & $97.97$ & $97.52$ & $97.17$ & $89.20$ & $58.11$\tabularnewline
\textit{ DGL2 $\lambda =$ 0.001 $\phi =$ 0.05} & $98.20$ & $97.88$ & $97.28$ & $89.41$ & $58.29$\tabularnewline
\textit{ DGL2 $\lambda =$ 0.001 $\phi =$ 0.1} & $98.26$ & $97.90$ & $97.43$ & $89.71$ & $58.21$\tabularnewline
\textit{ DGL2 $\lambda =$ 0.01 $\phi =$ 0.01} & $98.20$ & $97.92$ & $97.55$ & $91.95$ & $70.65$\tabularnewline
\textit{ DGL2 $\lambda =$ 0.01 $\phi =$ 0.05} & $98.31$ & $98.02$ & $97.69$ & $91.41$ & $66.94$\tabularnewline
\textit{ DGL2 $\lambda =$ 0.01 $\phi =$ 0.1} & $98.30$ & $98.00$ & $97.58$ & $91.21$ & $65.30$\tabularnewline
\textit{ L1 $\lambda =$ 0.0001} & $98.15$ & $97.84$ & $97.44$ & $89.63$ & $58.24$\tabularnewline
\textit{ L1 $\lambda =$ 0.001} & $98.41$ & $98.12$ & $97.68$ & $91.93$ & $68.21$\tabularnewline
\textit{ L1 $\lambda =$ 0.01} & $97.73$ & $97.54$ & $97.37$ & $95.01$ & $89.36$\tabularnewline
\textit{ L1 $\lambda =$ 0.1} & $93.90$ & $93.61$ & $93.40$ & $90.66$ & $85.01$\tabularnewline
\textit{ L2 $\lambda =$ 0.0001} & $98.00$ & $97.62$ & $97.16$ & $88.81$ & $56.66$\tabularnewline
\textit{ L2 $\lambda =$ 0.001} & $97.88$ & $97.49$ & $97.09$ & $88.85$ & $57.26$\tabularnewline
\textit{ L2 $\lambda =$ 0.01} & $98.45$ & $98.22$ & $98.02$ & $93.85$ & $78.97$\tabularnewline
\textit{ L2 $\lambda =$ 0.1} & $98.12$ & $98.02$ & $97.84$ & $95.29$ & $88.35$\tabularnewline
\textit{ Rect} & $97.99$ & $97.59$ & $97.12$ & $89.21$ & $58.22$\tabularnewline
\textit{MT-DGL2 $\lambda =$ 0.001 $\phi =$ 0.01 } & $98.38$ & $97.75$ & $96.73$ & $58.65$ & $10.89$\tabularnewline
\textit{MT-DGL2 $\lambda =$ 0.001 $\phi =$ 0.05 } & $98.50$ & $97.73$ & $96.75$ & $58.44$ & $10.43$\tabularnewline
\textit{MT-DGL2 $\lambda =$ 0.001 $\phi =$ 0.1 } & $98.35$ & $97.66$ & $96.42$ & $56.42$ & $10.20$\tabularnewline
\textit{MT-DGL2 $\lambda =$ 0.01 $\phi =$ 0.01 } & $98.57$ & $98.11$ & $97.62$ & $81.13$ & $29.40$\tabularnewline
\textit{MT-DGL2 $\lambda =$ 0.01 $\phi =$ 0.05 } & $98.65$ & $98.16$ & $97.62$ & $83.11$ & $31.86$\tabularnewline
\textit{MT-DGL2 $\lambda =$ 0.01 $\phi =$ 0.1 } & $98.59$ & $98.25$ & $97.73$ & $83.52$ & $33.65$\tabularnewline
\textit{MT-DGL2 $\lambda =$ 0.1 $\phi =$ 0.01 } & $98.85$ & $98.77$ & $98.59$ & $96.03$ & $85.77$\tabularnewline
\textit{MT-DGL2 $\lambda =$ 0.1 $\phi =$ 0.05 } & $98.76$ & $98.57$ & $98.37$ & $95.17$ & $81.90$\tabularnewline
\textit{MT-DGL2 $\lambda =$ 0.1 $\phi =$ 0.1 } & $98.70$ & $98.53$ & $98.39$ & $95.22$ & $83.77$\tabularnewline
\textit{MT-DGL1 $\lambda =$ 0.001 $\phi =$ 0.01 } & $98.58$ & $98.32$ & $97.93$ & $87.90$ & $46.80$\tabularnewline
\textit{MT-DGL1 $\lambda =$ 0.001 $\phi =$ 0.05 } & $98.77$ & $98.67$ & $98.50$ & $96.32$ & $89.44$\tabularnewline
\textit{MT-DGL1 $\lambda =$ 0.001 $\phi =$ 0.1 } & $98.63$ & $98.48$ & $98.23$ & $93.22$ & $76.42$\tabularnewline
\textit{MT-DGL1 $\lambda =$ 0.01 $\phi =$ 0.01 } & $98.82$ & $98.71$ & $98.47$ & $96.00$ & $86.98$\tabularnewline
\textit{MT-DGL1 $\lambda =$ 0.01 $\phi =$ 0.05 } & $98.91$ & $98.82$ & $98.73$ & $97.60$ & $94.22$\tabularnewline
\textit{MT-DGL1 $\lambda =$ 0.01 $\phi =$ 0.1 } & $98.99$ & $98.95$ & $98.84$ & $97.84$ & $95.56$\tabularnewline
\textit{MT-DGL1 $\lambda =$ 0.1 $\phi =$ 0.01 } & $98.63$ & $98.54$ & $98.46$ & $97.42$ & $95.18$\tabularnewline
\textit{MT-DGL1 $\lambda =$ 0.1 $\phi =$ 0.05 } & $98.99$ & $98.96$ & $98.92$ & $98.42$ & $97.39$\tabularnewline
\textit{MT-DGL1 $\lambda =$ 0.1 $\phi =$ 0.1 } & $\mathbf{99.03}$ & $\mathbf{99.00}$ & $\mathbf{98.98}$ & $\mathbf{98.55}$ & $\mathbf{97.77}$\tabularnewline
\textit{MT } & $98.00$ & $94.67$ & $87.73$ & $5.62$ & $1.15$\tabularnewline
\hline
\end{tabular}
}
\end{table}

\begin{table}[!h]
\caption{Comparative results where all models are attacked by the DataGrad-L1 regularized multi-task rectifier network, \emph{MT-DGL1}, using Laplacian ($\reg_1$) adversarial noise driven by the target task (digit recognition).}
\label{mt_dgl1_attack}
\centering
\resizebox{12.5cm}{!} {%
\begin{tabular}{lrrrrr}
\multicolumn{1}{l}{\begin{tabular}[x]{@{}c@{}}\textbf{Model}\\\end{tabular}}&\multicolumn{1}{c}{\begin{tabular}[x]{@{}c@{}}\textbf{$\phi$ = 0.0}\\\end{tabular}}&\multicolumn{1}{c}{\begin{tabular}[x]{@{}c@{}}\textbf{$\phi$ = 0.005}\\\end{tabular}}&\multicolumn{1}{c}{\begin{tabular}[x]{@{}c@{}}\textbf{$\phi$ = 0.01}\\\end{tabular}}&\multicolumn{1}{c}{\begin{tabular}[x]{@{}c@{}}\textbf{$\phi$ = 0.05}\\\end{tabular}}&\multicolumn{1}{c}{\begin{tabular}[x]{@{}c@{}}\textbf{$\phi$ = 0.1}\\\end{tabular}}\tabularnewline
\hline
\textit{ DGL1 $\lambda =$ 0.0001 $\phi =$ 0.01} & $98.25$ & $98.10$ & $97.94$ & $95.63$ & $89.74$\tabularnewline
\textit{ DGL1 $\lambda =$ 0.0001 $\phi =$ 0.05} & $98.57$ & $98.43$ & $98.27$ & $96.89$ & $92.92$\tabularnewline
\textit{ DGL1 $\lambda =$ 0.0001 $\phi =$ 0.1} & $98.47$ & $98.33$ & $98.24$ & $96.70$ & $92.79$\tabularnewline
\textit{ DGL1 $\lambda =$ 0.001 $\phi =$ 0.01} & $98.03$ & $97.91$ & $97.66$ & $95.47$ & $89.90$\tabularnewline
\textit{ DGL1 $\lambda =$ 0.001 $\phi =$ 0.05} & $98.70$ & $98.57$ & $98.42$ & $97.01$ & $93.27$\tabularnewline
\textit{ DGL1 $\lambda =$ 0.001 $\phi =$ 0.1} & $98.62$ & $98.50$ & $98.41$ & $96.93$ & $93.22$\tabularnewline
\textit{ DGL1 $\lambda =$ 0.01 $\phi =$ 0.01} & $98.36$ & $98.20$ & $98.07$ & $96.01$ & $91.09$\tabularnewline
\textit{ DGL1 $\lambda =$ 0.01 $\phi =$ 0.05} & $98.62$ & $98.52$ & $98.49$ & $96.88$ & $93.56$\tabularnewline
\textit{ DGL1 $\lambda =$ 0.01 $\phi =$ 0.1} & $\mathbf{98.83}$ & $\mathbf{98.77}$ & $\mathbf{98.68}$ & $\mathbf{97.72}$ & $\mathbf{95.26}$\tabularnewline
\textit{ DGL2 $\lambda =$ 0.0001 $\phi =$ 0.01} & $97.93$ & $97.79$ & $97.57$ & $95.33$ & $89.18$\tabularnewline
\textit{ DGL2 $\lambda =$ 0.0001 $\phi =$ 0.05} & $98.29$ & $98.15$ & $97.91$ & $95.69$ & $90.02$\tabularnewline
\textit{ DGL2 $\lambda =$ 0.0001 $\phi =$ 0.1} & $98.02$ & $97.85$ & $97.67$ & $95.22$ & $89.50$\tabularnewline
\textit{ DGL2 $\lambda =$ 0.001 $\phi =$ 0.01} & $97.97$ & $97.73$ & $97.52$ & $95.12$ & $89.20$\tabularnewline
\textit{ DGL2 $\lambda =$ 0.001 $\phi =$ 0.05} & $98.20$ & $98.06$ & $97.78$ & $95.31$ & $89.61$\tabularnewline
\textit{ DGL2 $\lambda =$ 0.001 $\phi =$ 0.1} & $98.26$ & $98.11$ & $98.00$ & $95.79$ & $90.29$\tabularnewline
\textit{ DGL2 $\lambda =$ 0.01 $\phi =$ 0.01} & $98.20$ & $98.02$ & $97.84$ & $95.81$ & $90.29$\tabularnewline
\textit{ DGL2 $\lambda =$ 0.01 $\phi =$ 0.05} & $98.31$ & $98.14$ & $97.99$ & $95.82$ & $90.24$\tabularnewline
\textit{ DGL2 $\lambda =$ 0.01 $\phi =$ 0.1} & $98.30$ & $98.14$ & $97.93$ & $95.61$ & $89.86$\tabularnewline
\textit{ L1 $\lambda =$ 0.0001} & $98.15$ & $98.01$ & $97.89$ & $95.51$ & $90.19$\tabularnewline
\textit{ L1 $\lambda =$ 0.001} & $98.41$ & $98.26$ & $98.01$ & $95.93$ & $90.82$\tabularnewline
\textit{ L1 $\lambda =$ 0.01} & $97.73$ & $97.50$ & $97.36$ & $95.25$ & $90.77$\tabularnewline
\textit{ L1 $\lambda =$ 0.1} & $93.90$ & $93.73$ & $93.53$ & $91.61$ & $86.95$\tabularnewline
\textit{ L2 $\lambda =$ 0.0001} & $98.00$ & $97.85$ & $97.66$ & $95.34$ & $89.38$\tabularnewline
\textit{ L2 $\lambda =$ 0.001} & $97.88$ & $97.71$ & $97.51$ & $95.02$ & $89.23$\tabularnewline
\textit{ L2 $\lambda =$ 0.01} & $98.45$ & $98.30$ & $98.14$ & $96.19$ & $91.15$\tabularnewline
\textit{ L2 $\lambda =$ 0.1} & $98.12$ & $98.04$ & $97.92$ & $96.12$ & $91.69$\tabularnewline
\textit{ Rect} & $97.99$ & $97.83$ & $97.63$ & $95.27$ & $89.83$\tabularnewline
\textit{MT-DGL2 $\lambda =$ 0.001 $\phi =$ 0.01 } & $98.38$ & $98.14$ & $97.90$ & $93.12$ & $80.36$\tabularnewline
\textit{MT-DGL2 $\lambda =$ 0.001 $\phi =$ 0.05 } & $98.50$ & $98.26$ & $97.96$ & $93.43$ & $80.28$\tabularnewline
\textit{MT-DGL2 $\lambda =$ 0.001 $\phi =$ 0.1 } & $98.35$ & $98.11$ & $97.86$ & $93.00$ & $79.57$\tabularnewline
\textit{MT-DGL2 $\lambda =$ 0.01 $\phi =$ 0.01 } & $98.57$ & $98.35$ & $98.08$ & $93.71$ & $81.96$\tabularnewline
\textit{MT-DGL2 $\lambda =$ 0.01 $\phi =$ 0.05 } & $98.65$ & $98.38$ & $98.05$ & $93.77$ & $82.71$\tabularnewline
\textit{MT-DGL2 $\lambda =$ 0.01 $\phi =$ 0.1 } & $98.59$ & $98.39$ & $98.16$ & $93.88$ & $82.13$\tabularnewline
\textit{MT-DGL2 $\lambda =$ 0.1 $\phi =$ 0.01 } & $98.85$ & $98.73$ & $98.53$ & $95.33$ & $85.69$\tabularnewline
\textit{MT-DGL2 $\lambda =$ 0.1 $\phi =$ 0.05 } & $98.76$ & $98.53$ & $98.32$ & $94.72$ & $84.69$\tabularnewline
\textit{MT-DGL2 $\lambda =$ 0.1 $\phi =$ 0.1 } & $98.70$ & $98.53$ & $98.38$ & $94.66$ & $84.27$\tabularnewline
\textit{MT-DGL1 $\lambda =$ 0.001 $\phi =$ 0.01 } & $98.58$ & $98.45$ & $98.21$ & $93.97$ & $82.65$\tabularnewline
\textit{MT-DGL1 $\lambda =$ 0.001 $\phi =$ 0.05 } & $98.77$ & $98.62$ & $98.44$ & $95.40$ & $86.81$\tabularnewline
\textit{MT-DGL1 $\lambda =$ 0.001 $\phi =$ 0.1 } & $98.63$ & $98.46$ & $98.28$ & $94.07$ & $84.26$\tabularnewline
\textit{MT-DGL1 $\lambda =$ 0.01 $\phi =$ 0.01 } & $98.82$ & $98.64$ & $98.33$ & $94.84$ & $85.10$\tabularnewline
\textit{MT-DGL1 $\lambda =$ 0.01 $\phi =$ 0.05 } & $98.91$ & $98.75$ & $98.55$ & $95.86$ & $88.21$\tabularnewline
\textit{MT-DGL1 $\lambda =$ 0.01 $\phi =$ 0.1 } & $98.99$ & $98.55$ & $97.96$ & $93.23$ & $86.97$\tabularnewline
\textit{MT-DGL1 $\lambda =$ 0.1 $\phi =$ 0.01 } & $98.63$ & $98.51$ & $98.35$ & $96.64$ & $93.01$\tabularnewline
\textit{MT-DGL1 $\lambda =$ 0.1 $\phi =$ 0.05 } & $98.99$ & $98.95$ & $98.85$ & $98.05$ & $95.82$\tabularnewline
\textit{MT-DGL1 $\lambda =$ 0.1 $\phi =$ 0.1 } & $\mathbf{99.03}$ & $\mathbf{99.00}$ & $\mathbf{98.88}$ & $\mathbf{98.12}$ & $\mathbf{96.71}$\tabularnewline
\textit{MT } & $98.00$ & $97.68$ & $97.40$ & $92.48$ & $79.59$\tabularnewline
\hline
\end{tabular}
}
\end{table}

\begin{table}[!h]
\caption{Comparative results where all models are attacked by the DataGrad-L2 regularized multi-task rectifier network, \emph{MT-DGL2}, using Laplacian ($\reg_1$) adversarial noise driven by the target task (digit recognition).}
\label{mt_dgl2_attack}
\centering
\resizebox{12.5cm}{!} {%
\begin{tabular}{lrrrrr}
\multicolumn{1}{l}{\begin{tabular}[x]{@{}c@{}}\textbf{Model}\\\end{tabular}}&\multicolumn{1}{c}{\begin{tabular}[x]{@{}c@{}}\textbf{$\phi$ = 0.0}\\\end{tabular}}&\multicolumn{1}{c}{\begin{tabular}[x]{@{}c@{}}\textbf{$\phi$ = 0.005}\\\end{tabular}}&\multicolumn{1}{c}{\begin{tabular}[x]{@{}c@{}}\textbf{$\phi$ = 0.01}\\\end{tabular}}&\multicolumn{1}{c}{\begin{tabular}[x]{@{}c@{}}\textbf{$\phi$ = 0.05}\\\end{tabular}}&\multicolumn{1}{c}{\begin{tabular}[x]{@{}c@{}}\textbf{$\phi$ = 0.1}\\\end{tabular}}\tabularnewline
\hline
\textit{ DGL1 $\lambda =$ 0.0001 $\phi =$ 0.01} & $98.25$ & $98.01$ & $97.78$ & $93.31$ & $78.21$\tabularnewline
\textit{ DGL1 $\lambda =$ 0.0001 $\phi =$ 0.05} & $98.57$ & $98.34$ & $98.21$ & $95.82$ & $88.17$\tabularnewline
\textit{ DGL1 $\lambda =$ 0.0001 $\phi =$ 0.1} & $98.47$ & $98.28$ & $98.16$ & $95.72$ & $88.42$\tabularnewline
\textit{ DGL1 $\lambda =$ 0.001 $\phi =$ 0.01} & $98.03$ & $97.81$ & $97.48$ & $93.29$ & $79.74$\tabularnewline
\textit{ DGL1 $\lambda =$ 0.001 $\phi =$ 0.05} & $98.70$ & $98.56$ & $98.38$ & $96.25$ & $89.84$\tabularnewline
\textit{ DGL1 $\lambda =$ 0.001 $\phi =$ 0.1} & $98.62$ & $98.48$ & $98.37$ & $96.32$ & $90.73$\tabularnewline
\textit{ DGL1 $\lambda =$ 0.01 $\phi =$ 0.01} & $98.36$ & $98.16$ & $97.91$ & $94.65$ & $84.87$\tabularnewline
\textit{ DGL1 $\lambda =$ 0.01 $\phi =$ 0.05} & $98.62$ & $98.50$ & $98.44$ & $96.49$ & $91.22$\tabularnewline
\textit{ DGL1 $\lambda =$ 0.01 $\phi =$ 0.1} & $\mathbf{98.83}$ & $\mathbf{98.77}$ & $\mathbf{98.67}$ & $\mathbf{97.66}$ & $\mathbf{94.71}$\tabularnewline
\textit{ DGL2 $\lambda =$ 0.0001 $\phi =$ 0.01} & $97.93$ & $97.68$ & $97.34$ & $92.68$ & $77.64$\tabularnewline
\textit{ DGL2 $\lambda =$ 0.0001 $\phi =$ 0.05} & $98.29$ & $98.07$ & $97.68$ & $93.07$ & $78.21$\tabularnewline
\textit{ DGL2 $\lambda =$ 0.0001 $\phi =$ 0.1} & $98.02$ & $97.76$ & $97.39$ & $92.76$ & $78.23$\tabularnewline
\textit{ DGL2 $\lambda =$ 0.001 $\phi =$ 0.01} & $97.97$ & $97.64$ & $97.38$ & $92.88$ & $77.47$\tabularnewline
\textit{ DGL2 $\lambda =$ 0.001 $\phi =$ 0.05} & $98.20$ & $97.97$ & $97.51$ & $92.93$ & $78.46$\tabularnewline
\textit{ DGL2 $\lambda =$ 0.001 $\phi =$ 0.1} & $98.26$ & $98.05$ & $97.72$ & $93.33$ & $78.70$\tabularnewline
\textit{ DGL2 $\lambda =$ 0.01 $\phi =$ 0.01} & $98.20$ & $97.94$ & $97.71$ & $93.61$ & $80.60$\tabularnewline
\textit{ DGL2 $\lambda =$ 0.01 $\phi =$ 0.05} & $98.31$ & $98.10$ & $97.79$ & $93.53$ & $80.19$\tabularnewline
\textit{ DGL2 $\lambda =$ 0.01 $\phi =$ 0.1} & $98.30$ & $98.03$ & $97.79$ & $93.31$ & $79.36$\tabularnewline
\textit{ L1 $\lambda =$ 0.0001} & $98.15$ & $97.94$ & $97.73$ & $93.21$ & $78.52$\tabularnewline
\textit{ L1 $\lambda =$ 0.001} & $98.41$ & $98.14$ & $97.83$ & $94.35$ & $81.61$\tabularnewline
\textit{ L1 $\lambda =$ 0.01} & $97.73$ & $97.46$ & $97.31$ & $94.35$ & $86.50$\tabularnewline
\textit{ L1 $\lambda =$ 0.1} & $93.90$ & $93.59$ & $93.41$ & $90.34$ & $83.60$\tabularnewline
\textit{ L2 $\lambda =$ 0.0001} & $98.00$ & $97.75$ & $97.38$ & $92.44$ & $77.56$\tabularnewline
\textit{ L2 $\lambda =$ 0.001} & $97.88$ & $97.57$ & $97.28$ & $92.46$ & $77.88$\tabularnewline
\textit{ L2 $\lambda =$ 0.01} & $98.45$ & $98.24$ & $98.07$ & $94.72$ & $83.30$\tabularnewline
\textit{ L2 $\lambda =$ 0.1} & $98.12$ & $98.01$ & $97.86$ & $95.11$ & $86.89$\tabularnewline
\textit{ Rect} & $97.99$ & $97.71$ & $97.43$ & $92.92$ & $78.48$\tabularnewline
\textit{MT-DGL2 $\lambda =$ 0.001 $\phi =$ 0.01 } & $98.38$ & $97.93$ & $97.33$ & $80.38$ & $43.44$\tabularnewline
\textit{MT-DGL2 $\lambda =$ 0.001 $\phi =$ 0.05 } & $98.50$ & $98.02$ & $97.31$ & $80.83$ & $43.89$\tabularnewline
\textit{MT-DGL2 $\lambda =$ 0.001 $\phi =$ 0.1 } & $98.35$ & $97.87$ & $97.10$ & $80.19$ & $43.70$\tabularnewline
\textit{MT-DGL2 $\lambda =$ 0.01 $\phi =$ 0.01 } & $98.57$ & $98.15$ & $97.61$ & $83.33$ & $47.05$\tabularnewline
\textit{MT-DGL2 $\lambda =$ 0.01 $\phi =$ 0.05 } & $98.65$ & $98.14$ & $97.63$ & $84.72$ & $49.28$\tabularnewline
\textit{MT-DGL2 $\lambda =$ 0.01 $\phi =$ 0.1 } & $98.59$ & $98.25$ & $97.66$ & $84.46$ & $49.16$\tabularnewline
\textit{MT-DGL2 $\lambda =$ 0.1 $\phi =$ 0.01 } & $98.85$ & $98.36$ & $97.63$ & $73.98$ & $34.38$\tabularnewline
\textit{MT-DGL2 $\lambda =$ 0.1 $\phi =$ 0.05 } & $98.76$ & $98.42$ & $98.04$ & $89.93$ & $61.15$\tabularnewline
\textit{MT-DGL2 $\lambda =$ 0.1 $\phi =$ 0.1 } & $98.70$ & $98.45$ & $98.16$ & $90.77$ & $63.43$\tabularnewline
\textit{MT-DGL1 $\lambda =$ 0.001 $\phi =$ 0.01 } & $98.58$ & $98.27$ & $97.79$ & $85.03$ & $50.55$\tabularnewline
\textit{MT-DGL1 $\lambda =$ 0.001 $\phi =$ 0.05 } & $98.77$ & $98.56$ & $98.34$ & $92.90$ & $74.70$\tabularnewline
\textit{MT-DGL1 $\lambda =$ 0.001 $\phi =$ 0.1 } & $98.63$ & $98.36$ & $97.95$ & $88.59$ & $65.79$\tabularnewline
\textit{MT-DGL1 $\lambda =$ 0.01 $\phi =$ 0.01 } & $98.82$ & $98.52$ & $98.23$ & $92.12$ & $70.56$\tabularnewline
\textit{MT-DGL1 $\lambda =$ 0.01 $\phi =$ 0.05 } & $98.91$ & $98.76$ & $98.54$ & $95.67$ & $85.40$\tabularnewline
\textit{MT-DGL1 $\lambda =$ 0.01 $\phi =$ 0.1 } & $98.99$ & $98.89$ & $98.73$ & $95.84$ & $89.01$\tabularnewline
\textit{MT-DGL1 $\lambda =$ 0.1 $\phi =$ 0.01 } & $98.63$ & $98.49$ & $98.30$ & $96.38$ & $91.47$\tabularnewline
\textit{MT-DGL1 $\lambda =$ 0.1 $\phi =$ 0.05 } & $98.99$ & $98.94$ & $98.88$ & $97.93$ & $95.46$\tabularnewline
\textit{MT-DGL1 $\lambda =$ 0.1 $\phi =$ 0.1 } & $\mathbf{99.03}$ & $\mathbf{99.00}$ & $\mathbf{98.90}$ & $\mathbf{98.11}$ & $\mathbf{96.59}$\tabularnewline
\textit{MT } & $98.00$ & $97.43$ & $96.80$ & $82.60$ & $49.99$\tabularnewline

\hline
\end{tabular}
}
\end{table}

\end{document}